\documentclass[11pt]{article}
\usepackage{tabularx}  
\usepackage{array}     
\usepackage{booktabs}  
\usepackage{enumitem}
\usepackage[final]{latex/acl}
\usepackage{tikz}
\usepackage{pgfplots}
\usepgfplotslibrary{groupplots}
\pgfplotsset{compat=1.17}
\usepackage{booktabs}
\usepackage{amsmath}
\usepackage{cleveref}
\usepackage{siunitx}
\usepackage{ragged2e}
\usepackage{tcolorbox}
\tcbuselibrary{skins}
\usepackage{times}
\usepackage{latexsym}
\usepackage{multirow} 
\usepackage[T1]{fontenc}

\usepackage[utf8]{inputenc}
\usepackage{subcaption}
\usepackage{microtype}
\usepackage{algorithm}
\usepackage{algpseudocode}

\usepackage{inconsolata}

\usepackage{graphicx}
\usepackage{siunitx}
\usepackage{enumitem}

\title{\textsc{MTR-Suite}: A Framework for Evaluating and Synthesizing Conversational Retrieval Benchmarks}

\author{
  \textbf{Junhao Ruan\textsuperscript{1,2}\thanks{Work done during internship at Meituan.}},
  \textbf{Abudukeyumu Abudula\textsuperscript{1}},
  \textbf{Bei Li\textsuperscript{2}\thanks{Corresponding authors.}},
  \textbf{Yongjing Yin\textsuperscript{2}},
\\
  \textbf{Xinyu Liu\textsuperscript{1}},
  \textbf{Kechen Jiao\textsuperscript{4}},
  \textbf{Xin Chen\textsuperscript{2}},
  \textbf{Jingang Wang\textsuperscript{2}},
\\
  \textbf{Xunliang Cai\textsuperscript{2}},
  \textbf{Tong Xiao\textsuperscript{1,3}\footnotemark[2]},
  \textbf{Jingbo Zhu\textsuperscript{1,3}}
\\
\\
  \textsuperscript{1}School of Computer Science and Engineering, Northeastern University, Shenyang 110819, China
\\
  \textsuperscript{2}Meituan Inc.,
  \textsuperscript{3}NiuTrans Research, Shenyang, China
\\
  \textsuperscript{4}Tsinghua University, Beijing, China
\\
  \small{\texttt{rangehow@outlook.com}}
}

\begin{document}
\maketitle

\begin{abstract}

Accurate evaluation of conversational retrieval is pivotal for advancing Retrieval-Augmented Generation (RAG) systems. However, existing conversational retrieval benchmarks suffer from costly, sparse human annotation or rigid, unnatural automated heuristics. To address these challenges, we introduce \textsc{MTR-Suite}, a unified framework for auditing, synthesizing, and benchmarking retrieval. It features: (1) \textsc{MTR-Eval}, an LLM-based auditor quantifying alignment gaps in previous benchmarks; (2) \textsc{MTR-Pipeline}, a multi-agent system using greedy traversal clustering to generate high-fidelity dialogues at 1/400th human cost; and (3) \textsc{MTR-Bench}, a rigorous general-domain benchmark. \textsc{MTR-Bench} mimics production-style challenges (hard topic switching, verbosity), offering superior discriminative power. We make our code and data publicly available to facilitate future research.\footnote{\url{https://github.com/rangehow/mtr-suite}}

\end{abstract}
\begin{figure}[t]
    \centering
    \includegraphics[width=\linewidth]{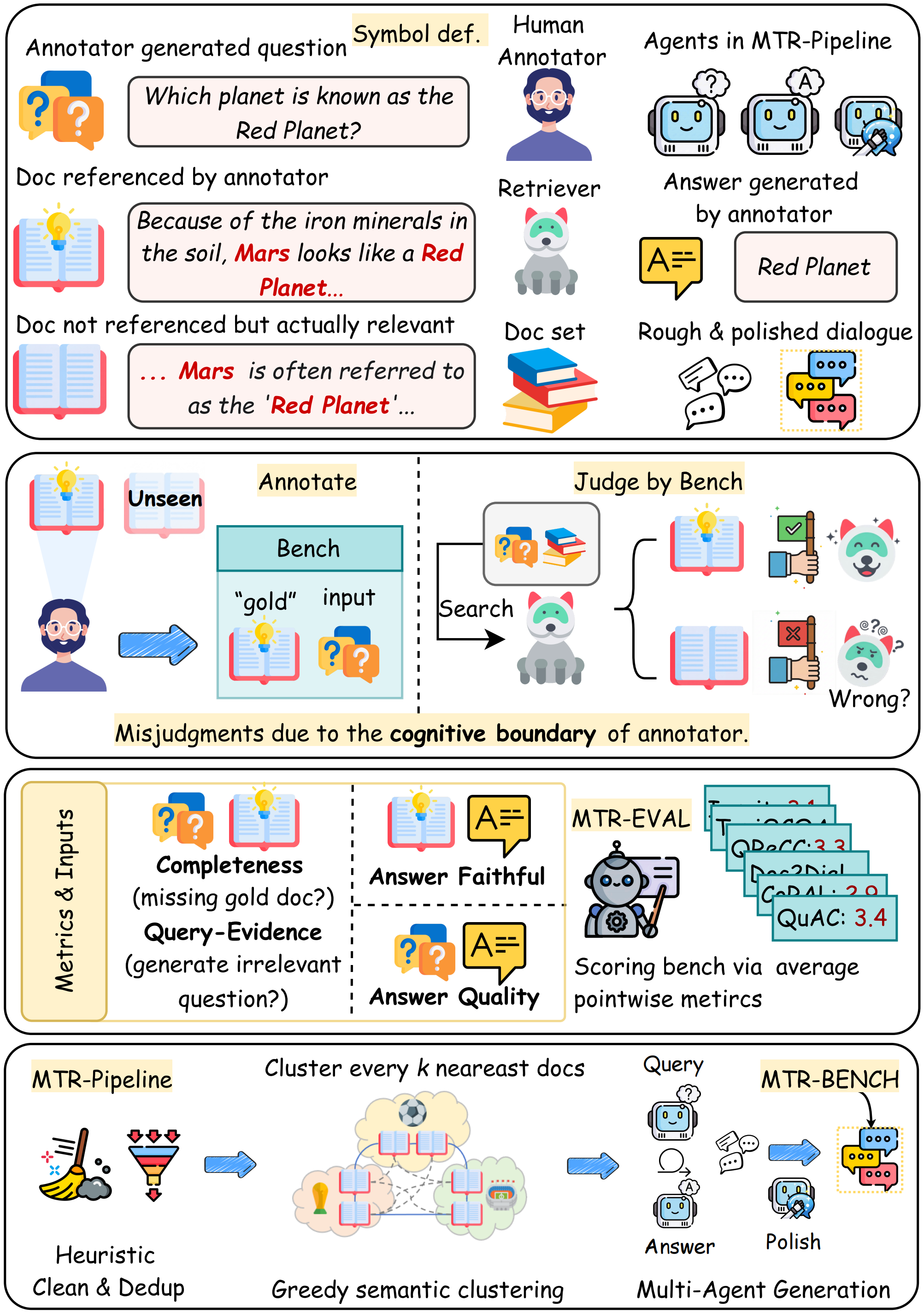}
    \caption{Top: Symbol definitions. Second Row: Illustration of the human \textit{cognitive boundary}. Third Row: The \textsc{MTR-Eval} auditing process. Bottom: The \textsc{MTR-Pipeline}, which synthesize high-quality benchmarks.}
    \label{fig:intro}
\end{figure}

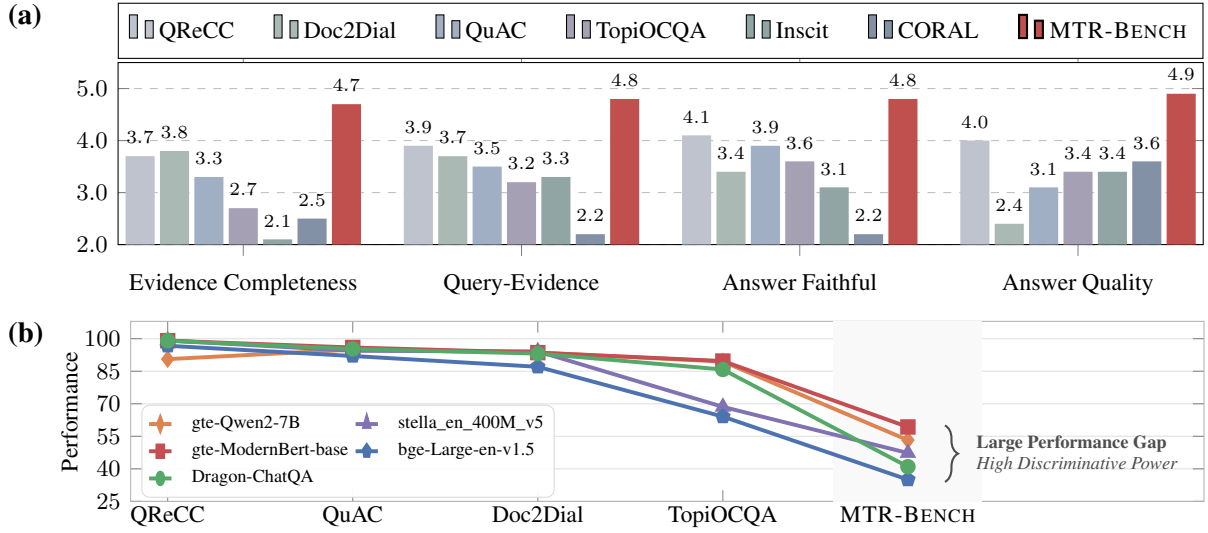
\begin{figure*}[t]
    \centering
    
    \begin{minipage}{\textwidth}
        \textbf{(a)} \\[-1em] 
        \resizebox{\textwidth}{!}{
\definecolor{color1}{RGB}{190, 195, 205}   
\definecolor{color2}{RGB}{170, 185, 180}   
\definecolor{color3}{RGB}{160, 175, 195}   
\definecolor{color4}{RGB}{165, 160, 180}   
\definecolor{color5}{RGB}{145, 165, 165}   
\definecolor{color6}{RGB}{130, 145, 165}   

\definecolor{color7}{RGB}{192, 80, 77}     

\begin{tikzpicture}
\begin{axis}[
    ybar,
    ymin=2,
    ymax=5.5, 
    width=16cm,
    height=4cm,
    bar width=11pt, 
    symbolic x coords={Evidence Completeness, Query-Evidence , Answer Faithful , Answer Quality},
    xtick=data,
    xticklabel style={rotate=0, font=\small, yshift=-3pt},
    tick label style={font=\small},
    yticklabel style={
        font=\small,
        /pgf/number format/.cd,
        fixed,
        fixed zerofill,
        precision=1
    },
    legend style={
        at={(0.5, 1.02)}, 
        anchor=south,     
        legend columns=-1, 
        /tikz/every even column/.append style={column sep=0.45cm}, 
        font=\footnotesize, 
        draw=black,       
        fill=white,
        inner sep=4pt,
    },
    ymajorgrids=true,
    grid style=dashed,
    enlarge x limits=0.15, 
    nodes near coords,
    nodes near coords style={font=\scriptsize, yshift=1pt}, 
    /pgf/number format/.cd,
    fixed,
    fixed zerofill,
    precision=1
]

\addplot[fill=color1, draw=none] coordinates {(Evidence Completeness, 3.7) (Query-Evidence , 3.9) (Answer Faithful , 4.1) (Answer Quality, 4.0)};
\addplot[fill=color2, draw=none] coordinates {(Evidence Completeness, 3.8) (Query-Evidence , 3.7) (Answer Faithful , 3.4) (Answer Quality, 2.4)};
\addplot[fill=color3, draw=none] coordinates {(Evidence Completeness, 3.3) (Query-Evidence , 3.5) (Answer Faithful , 3.9) (Answer Quality, 3.1)};
\addplot[fill=color4, draw=none] coordinates {(Evidence Completeness, 2.7) (Query-Evidence , 3.2) (Answer Faithful , 3.6) (Answer Quality, 3.4)};
\addplot[fill=color5, draw=none] coordinates {(Evidence Completeness, 2.1) (Query-Evidence , 3.3) (Answer Faithful , 3.1) (Answer Quality, 3.4)};
\addplot[fill=color6, draw=none] coordinates {(Evidence Completeness, 2.5) (Query-Evidence , 2.2) (Answer Faithful , 2.2) (Answer Quality, 3.6)};

\addplot[fill=color7, draw=none, thick] coordinates {(Evidence Completeness, 4.7) (Query-Evidence , 4.8) (Answer Faithful , 4.8) (Answer Quality, 4.9)};

\legend{QReCC, Doc2Dial, QuAC, TopiOCQA, Inscit, CORAL, \textsc{MTR-Bench}}

\end{axis}
\end{tikzpicture} 
        }
    \end{minipage}
    
    \vspace{0.5em} 
    
    \begin{minipage}{\textwidth}
        \textbf{(b)} \\[-1em]
        \resizebox{\textwidth}{!}{
            \pgfplotsset{compat=1.18}
\usetikzlibrary{patterns, positioning, decorations.pathreplacing, arrows.meta}
\definecolor{myblue}{RGB}{76, 114, 176}
\definecolor{myorange}{RGB}{221, 132, 82}
\definecolor{mygreen}{RGB}{85, 168, 104}
\definecolor{myred}{RGB}{196, 78, 82}
\definecolor{mypurple}{RGB}{129, 114, 179}

\begin{tikzpicture}
\begin{axis}[
    width=\linewidth,
    height=4cm,
    xmin=0.8, xmax=6.6,
    ymin=25, ymax=108,
    xtick={1,2,3,4,5},
    xticklabels={QReCC, QuAC, Doc2Dial, TopiOCQA, \textsc{MTR-Bench}},
    ytick={25, 40, 55, 70, 85, 100},
    tick label style={font=\small},
    xticklabel style={yshift=2pt},
    ylabel={Performance},
    ylabel style={font=\small, yshift=-5pt},
    ymajorgrids=true,
    xmajorgrids=false,
    grid style={line width=0.5pt, draw=gray!20},
    clip=false,
    legend style={
        at={(0.01,0.02)},
        anchor=south west,
        legend columns=2, 
        legend cell align=left,
        font=\scriptsize,
        inner sep=2pt,
        row sep=0pt,
        /tikz/every odd column/.append style={column sep=5pt},
        draw=gray!30,
        rounded corners=3pt,
        fill=white,
        fill opacity=0.95,
        text opacity=1
    },
    legend image post style={xscale=0.5},
    mark options={solid, scale=1.1},
    every axis plot/.append style={thick, line width=1.5pt},
    axis line style={draw=gray!60}
]

\fill[gray!5] (axis cs:4.6, 25) rectangle (axis cs:5.4, 108);

\addplot[color=myorange, mark=diamond*] coordinates { (1,90.58) (2,94.97) (3,93.58) (4,89.46) (5,53.23) };
\addlegendentry{gte-Qwen2-7B}

\addplot[color=mypurple, mark=triangle*] coordinates { (1,99.21) (2,94.41) (3,94.16) (4,68.58) (5,47.30) };
\addlegendentry{stella\_en\_400M\_v5}

\addplot[color=myred, mark=square*] coordinates { (1,99.10) (2,95.95) (3,93.58) (4,89.70) (5,59.31) };
\addlegendentry{gte-ModernBert-base}

\addplot[color=myblue, mark=pentagon*] coordinates { (1,96.74) (2,92.02) (3,87.05) (4,64.08) (5,35.04) };
\addlegendentry{bge-Large-en-v1.5}

\addplot[color=mygreen, mark=*] coordinates { (1,99.03) (2,95.24) (3,93.15) (4,85.80) (5,40.98) };
\addlegendentry{Dragon-ChatQA}


\draw[decorate, decoration={brace, amplitude=5pt, mirror}, thick, black!70] 
    (axis cs:5.2, 34) -- (axis cs:5.2, 60) 
    node [midway, xshift=8pt, anchor=west, align=left, font=\scriptsize] 
    (bracelabel)
    {\textbf{Large Performance Gap} \\ \textit{High Discriminative Power}};

\end{axis}
\end{tikzpicture}

        }
    \end{minipage}

    \caption{
        (a) Average quantitative analysis results of multiple powerful open-source LLMs on different datasets using our proposed \textsc{MTR-Eval} metrics. Detailed score heatmaps for each model across all benchmarks are provided in Appendix~\ref{app:previous_benchmark_heatmap}.
        (b) Recall@20 scores of popular open-source retrievers on various benchmarks. 
        Note the larger performance gap in \textsc{MTR-Bench}, highlighting its discriminative power compared to previous benchmarks.
    }
    \label{fig:quality_assessment}

    \phantomsubcaption \label{fig:quality_a}
    \phantomsubcaption \label{fig:quality_b}
    
\end{figure*}

\section{Introduction}

Retrieval-Augmented Generation (RAG) has emerged as the standard paradigm for grounding Large Language Models (LLMs) in verifiable, external knowledge \citep{RAGORIGIN, openai2023chatgpt, geminiteam2024gemini15unlockingmultimodal, deepseekai2024deepseekv3technicalreport}. By mitigating the hallucinations inherent to static parametric memory \citep{xu2025hallucinationinevitableinnatelimitation}, RAG extends the capabilities of models further amplified by tool use \citep{toolformer} and test-time scaling \citep{o1, r1}. However, the efficacy of any RAG system remains bounded by the performance of its retrieval module. If the retriever fails to locate the precise conversational context, the generation layer is rendered ineffective.

Consequently, evaluating the conversational retrieval stage is critical. Traditional manual annotation faces a fundamental \textit{cognitive boundary}~\citep{zobel1998reliable, buckley2004retrieval}. Annotators typically operate with a ``local view'', which formulating queries based solely on the specific document they are reading. They lack the global perspective to know if other documents in the corpus could also answer the query (potentially better). This leads to \textit{annotation sparsity}, where valid retrievals are penalized as false negatives. Furthermore, strict privacy constraints often make manual annotation prohibitive for high-value proprietary domains (e.g., finance, legal).

Ideally, automated synthesis should overcome these human limitations. However, prior attempts like CORAL \citep{coral} have inadvertently inherited similar challenges. By relying on static heuristics, such as directly converting Wikipedia section headers into questions, these methods operate within a methodological boundary that mirrors the human local view. They assume that a query generated from a specific document structure is uniquely aligned with that document. As our auditing suggests (Section \ref{sec:quality-assessment}), without global validation, this approach can result in alignment issues: generated queries may be ambiguous or better answered by unannotated documents elsewhere in the corpus. Thus, existing automation trades the high cost of humans for efficiency, but the challenge of annotation sparsity remains to be fully addressed.

To bridge this gap, we argue that high-fidelity benchmarking requires a transition from local heuristics to Global-Aware Automated Annotation.
We introduce \textsc{MTR-Suite} (Figure \ref{fig:intro}), a unified framework designed to audit, synthesize, and benchmark conversational retrieval systems. \textsc{MTR-Suite} provides a holistic solution through three integrated contributions:

\begin{itemize}[leftmargin=*]
    \item \textbf{Diagnostic Auditing (\textsc{MTR-Eval}):} We propose the first fine-grained evaluation method designed to scientifically quantify the quality of retrieval benchmarks. Our experiments reveal that while automated baselines suffer from linguistic degradation, even human-annotated datasets exhibit significant sparsity-induced noise. \textsc{MTR-Eval} correlates highly with human judgments, offering a reliable metric for dataset integrity (Figure \ref{fig:quality_a}).
    
    \item \textbf{Multi-Agent Synthesis (\textsc{MTR-Pipeline}):} We introduce a fully automated framework for constructing multi-turn retrieval datasets. Unlike rule-based predecessors, our pipeline utilizes a sophisticated multi-agent architecture driven by a greedy traversal clustering algorithm. This approach ensures query uniqueness and linguistic naturalness, achieving annotation quality that surpasses human standards at a fraction of the cost ($1/400$th).
    
    \item \textbf{A Rigorous Benchmark (\textsc{MTR-Bench}):} We open-source \textsc{MTR-Bench}, a general-domain dataset designed to stress-test modern retrievers. Distinguished by realistic conversational phenomena often ignored by previous datasets, including hard topic switching, long-context ambiguity, and production-style verbosity, it provides significantly higher discriminative power for evaluating mainstream retrieval modules (Figure \ref{fig:quality_b}).
\end{itemize}

\section{Related Work}
\label{sec:related_work}

The evolution of conversational search benchmarks reflects a continuous balancing act between annotation quality, scalability, and cost. We categorize prior efforts into three main paradigms.

\subsection{Manual and Hybrid Annotation}
Foundational datasets like CoQA~\citep{reddy-etal-2019-coqa} and QuAC~\citep{choi-etal-2018-quac} established the standard for multi-turn interactions, while subsequent works targeted specific domains, such as goal-oriented dialogues in Doc2Dial~\citep{feng-etal-2020-doc2dial} and topic shifts in TopiocQA~\citep{adlakha-etal-2022-topiocqa}.
To mitigate annotation costs, hybrid semi-synthetic approaches like TREC CAsT~\citep{dalton2020treccast2019conversational} and QReCC~\citep{qrecc} combine human rewrites with machine retrieval. While improving efficiency, they inherit the fundamental limitations of human oversight and struggle to scale effectively to private, large-scale corpora. Despite their status as gold standards, these datasets suffer from the human cognitive limitations discussed in introduction. Specifically, the lack of global corpus visibility leads to significant false negatives (retrievable evidence labeled as irrelevant) and hallucinated queries. As demonstrated in our case studies (see Appendix~\ref{app:case_study}), even these manually curated datasets exhibit considerable noise and sparsity. 

\subsection{Automated Benchmark Synthesis}
The emergence of LLMs has shifted focus toward fully automated construction. Notable examples like CORAL~\citep{coral} synthesize dialogues by leveraging the hierarchical structure of Wikipedia (e.g., turning section headings into questions). However, reliance on static document structures may restrict conversational naturalness, resulting in dialogues that mirror the source text's organization rather than dynamic user intent. \citet{lee2024multidocumentgroundedmultiturnsynthetic} introduced a multi-document synthesis pipeline. Their reliance on rigid rules to enforce specific question types can compromise conversational flow, leading to unnatural transitions such as illogical counter-arguments to correct responses. The unavailability of this dataset also precludes assessment of its quality.

\section{\textsc{MTR-Eval}: Auditing Benchmark Quality}
\label{sec:quality-assessment}

Formally, a multi-turn conversational retrieval benchmark consists of a document corpus $\mathcal{D} = \{d_1, d_2, \dots, d_N\}$, where each $d$ represents an individual document (or passage), and a set of evaluation instances. Each instance corresponds to a conversational turn $i$, represented as a tuple $(H_i, q_i, G_i)$. Here, $q_i$ is the user's current query, and $H_i = \{(q_1, a_1), \dots, (q_{i-1}, a_{i-1})\}$ denotes the conversation history. The set $G_i \subseteq \mathcal{D}$ contains the ground-truth documents annotated as relevant for answering $q_i$.

\subsection{The Quality Gap: Annotated vs. True Evidence}
The fundamental challenge in benchmarking is the discrepancy between the \textit{annotated} ground-truth set $G_i$ and the \textit{true} set of all supporting documents $\hat{G}_i$ existing in the corpus $\mathcal{D}$. A high-quality benchmark must minimize two specific error types:

\paragraph{Annotation Noise:} When $G_i \setminus \hat{G}_i \neq \emptyset$, the benchmark includes documents $d$ in the gold set that are actually irrelevant. This leads to model overestimation.
\paragraph{Annotation Sparsity
:} When $\hat{G}_i \setminus G_i \neq \emptyset$, valid supporting documents $d$ exist in the corpus but are missing from the annotation. This causes valid retrievals to be penalized as false negatives.

\subsection{The Evaluation Metrics}
To quantify these discrepancies without requiring exhaustive manual re-annotation, \textsc{MTR-Eval} employs an heterogeneous LLM-as-a-Judge approach across four fine-grained dimensions:

\paragraph{Query-Evidence Alignment:} 
This metric assesses whether a specific annotated document $d \in G_i$ actually contains the answer to $q_i$. The LLM is presented with the query and the gold document; if $d$ fails to support the query, it indicates \textit{Annotation Noise}, meaning the benchmark contains false positives.

\paragraph{Evidence Completeness:} 
Measuring \textit{Annotation Sparsity} is difficult because enumerating all relevant documents in a large corpus is intractable. We propose a proxy task: Discriminability Testing.
We present the LLM with a candidate pool containing the gold document $d_{gold} \in G_i$ and several retrieved hard negatives. The LLM is asked to identify the most relevant document. If the LLM consistently selects a non-gold document $d_{other} \notin G_i$ as being \textit{more} relevant than $d_{gold}$, it strongly suggests the existence of valid evidence that was missed during annotation. This metric penalizes benchmarks where the ``gold'' document is not uniquely or clearly the best match.

\paragraph{Answer-Evidence Faithfulness:} 
This evaluates hallucinations during annotation. We verify if the answer $a_i$ is fully grounded in its corresponding document set $G_{i}$.

\paragraph{Answer Quality:} 
Independent of evidence, this metric evaluates the linguistic quality of the response $a_i$ (e.g., coherence, helpfulness for $q_{i}$).

Collectively, these metrics act as a reliability coefficient for the dataset. A critical implication of our methodology is that achieving high performance with a retriever is only significant on benchmarks that possess high \textsc{MTR-Eval} scores. Conversely, high scores (e.g., Recall or mrr) on benchmarks with low \textsc{MTR-Eval} ratings hold limited value, as the results are likely conflated by annotation noise and sparsity.

\subsection{Validation}
\paragraph{Verification of Missing Evidence.} 
To verify the detected sparsity, we manually examined 300 sampled cases where the LLM identified a document as relevant despite its absence from the ground truth. Human review confirmed that 98\% of these were indeed valid gold(under 0.92 Fleiss' Kappa score) documents sufficient to fully answer the user's query, validating the existence of the cognitive boundary. More cases are available in appendix~\ref{app:case_study}.

\paragraph{Human Correlations.}
We conducted a parallel human annotation study. Critically, our method minimizes biases through a multi-LLM ensemble (reducing self-preference bias) and a pointwise scoring design (eliminating position bias). In multi-document scenarios like Discriminability Testing, document positions are further randomized. These design choices resulted in automated scores that align strongly with human judgments, achieving Pearson correlations of $0.82$ for Query-Evidence Relevance, $0.89$ for Answer Faithfulness, and $0.74$ for Answer Quality (all $p \ll 0.001$). This high alignment confirms that MTR-EVAL serves as a reliable proxy for human perception, decoupled from specific model preferences. Further details are provided in Appendix~\ref{appendix:human_corr}.

\section{\textsc{MTR-Pipeline}}
To generate high-quality data at scale, we propose \textsc{MTR-Pipeline}. We decompose the complex task of benchmark construction into three streamlined stages: Curation, Clustering, and Multi Agent Generation. For comprehensive engineering implementation details regarding recursive chunking guidelines, clustering case study, agent model selection, and cost estimation breakdowns, please refer to Appendix~\ref{app:mtr-pipeline}.

\subsection{Knowledge Base Curation}
The quality of a benchmark is strictly bounded by its source data. To ensure only high-quality segments serve as reference documents, we process the raw corpus through a streamlined pipeline. We first remove non-textual elements and apply recursive chunking to optimize context window usage, followed by MinHash-LSH~\cite{broder1997resemblance} to eliminate near-duplicate chunks. 
Finally, we employ a hybrid quality filter that utilizes an NVIDIA quality classifier for fluency and the FineWeb-EDU scorer~\citep{penedo2024finewebdatasetsdecantingweb} for educational value. We retain only those chunks that exhibit high information density and score as ``High Quality.''

\subsection{Greedy Clustering: Scalable Semantic Trajectories}

To simulate natural topic-switching behaviors, such as a user following a chain of hyperlinks, the pipeline must feed the LLM sequences of semantically related documents. However, conventional clustering algorithms like K-means or DBSCAN are ill-suited for this task, as they produce clusters of uncontrollable sizes that often exceed the LLM's context window. Furthermore, a naive approach of selecting neighbors for each document based on a similarity threshold is also problematic: it leads to high document overlap across different clusters, causing the LLM to encounter the same content repeatedly during training and evaluation, which introduces significant evaluation bias.

To address these limitations, we propose a Greedy Traversal Clustering strategy. While greedy nearest-neighbor search is a classical heuristic for the Traveling Salesperson Problem (TSP), its application as a clustering mechanism for ``soft topic switching'' is highly novel. Instead of static grouping, we repurpose this traditional logic to construct a single, continuous ``semantic path'' by iteratively selecting the nearest unvisited neighbor from a random starting node. This path is then segmented every $k$ nodes. 

This approach offers three distinct advantages: (1) it guarantees a fixed cluster size $k$ to maximize context utilization; (2) by ensuring each document is visited exactly once, it eliminates the redundancy and bias inherent in threshold-based methods; and (3) it creates a smooth semantic gradient that naturally mimics a user's browsing trajectory, providing a more realistic foundation for simulating fluid topic transitions than any conventional clustering method.

\subsection{Multi-Agent Dialogue Generation}
To simulate realistic conversational dynamics, we employ a three-agent system. Decomposing the generation process into specific roles enhances instruction adherence and data quality:

\paragraph{The Questioner (User Simulator):} 
Formulates queries based on the document cluster and conversation history. It decides when to switch topics or drill down into specific details.

\paragraph{The Responder (RAG Simulator):} 
Generates answers strictly grounded in the designated gold document. This agent mimics the behavior of an ideal RAG system.

\paragraph{The Polisher (Refiner):} 
Post-processes the dialogue to enhance linguistic naturalness, injecting conversational phenomena such as coreference (e.g., ``What about \textit{him}?'') and ellipsis to mimic human speech patterns.

\paragraph{Scalability and Cost Efficiency.} 
Beyond linguistic quality, a critical advantage of our automated framework is its economic viability. Based on current API pricing (e.g., DeepSeek-V3.2), we estimate the average cost per synthesized dialogue to be approximately \$0.005. This represents a dramatic reduction (roughly 1/400) compared to the \$1.50--\$2.00 per dialogue reported in crowd sourced benchmarks like Doc2Dial~\citep{feng-etal-2020-doc2dial}. This efficiency enables the scalable construction of large-scale, domain-specific datasets. 

\begin{table*}[ht]
\centering
\small
\begin{tabular}{lcccccccc}
\toprule
    \textbf{Benchmark} & \textbf{KC} & \textbf{RL} & \textbf{Docs/Query} & \textbf{Topic Switch} & \textbf{DL} & \textbf{Interaction Mode} & \textbf{CT} \\
\midrule
QuAC \citeyearpar{choi-etal-2018-quac} & $\le 2018 $ & 18.91 & 19.59 &  N/A  & 2079.64 & Human-Human &7.38 \\
Doc2Dial~\citeyearpar{feng-etal-2020-doc2dial} & $\le 2020 $ & 21.77 & 309.75 &  N/A  & \textbf{1730.16}$^\ddagger$& Human-Human & 6.52 \\
QReCC~\citeyearpar{qrecc} & 2019 & 27.16 & 14.97 & soft (part) & 2152.48 & Human-Human &4.36 \\

TopiOCQA \citeyearpar{adlakha-etal-2022-topiocqa} & 2020 & 9.09 & \textbf{\textasciitilde 20M} & soft & 395.03 & Human-Human &12.26 \\
INSCIT \citeyearpar{wu-etal-2023-inscit} & 2021 & 35.88 & \textbf{\textasciitilde 50M} &  N/A  & 493.06 & Human-Human &5.84 \\

CORAL \citeyearpar{coral} & $\le$ 2019 & \textbf{247.04}$^\dagger$ & \textasciitilde 200k &  soft  & $\textbf{1346.47}^\ddagger$ & Agent - Rule & $\textbf{8.1}^\dagger$\\

\textsc{MTR-Bench} & \textbf{2025} & $\textbf{86.90}^\dagger$ & \textbf{\textasciitilde 1M} & \textbf{soft \& hard} & $\textbf{1678.71}^\ddagger$ & \textbf{Simulated User - Agent} & $\textbf{8}^\dagger$ \\
\bottomrule
\end{tabular}

\caption{Comparison of \textsc{MTR-Bench} with previous benchmarks. Settings that more closely align with practical application scenarios are highlighted in \textbf{bold}. The abbreviations are as follows: KC (Knowledge Cutoff), RL (Response Length in tokens), and CT (Conversation Turns). The design choices for items marked with $\dagger$ are based on an analysis of large-scale, real user-agent dialogues from ShareGPT. The choice for the item marked with $\ddagger$ is informed by prior research from LlamaIndex. For KC, dates marked with `$\le$` were not explicitly specified in the original papers and are our estimations based on the paper release dates or their documented data sources.}
\label{tab:database_info}
\end{table*}

\begin{table}[h]
\centering
\small
\begin{tabular*}{\linewidth}{@{\extracolsep{\fill}}lrrr@{}}
\toprule
\textbf{Statistic} & \textbf{Dev} & \textbf{Test} & \textbf{Overall} \\
\midrule
\# Turns & 31,896 & 8,000 & 39,896 \\
\# Conversations & 3,987 & 1,000 & 4,987 \\
Tokens / Question & 15.32 & 15.35 & 15.33 \\
Tokens / Answer & 87.67 & 86.90 & 87.52 \\
Turns / Conversation & 8 & 8 & 8 \\
Topics / Conversation & 5.59 & 5.70 & 5.61 \\
\bottomrule
\end{tabular*}
\caption{Statistical overview of the MTR dataset. `Dev' refers to MTR-dev, which can be partitioned from the training set for model debugging and development, but is not used for evaluation in this paper. `Test' refers to the primary evaluation benchmark \textsc{MTR-Bench}. Note: we publicly release an improved 12-turn version synthesized by Qwen3.5 at our repository, which we recommend for future research.}
\label{tab:dataset-analysis}
\end{table}
\begin{figure}[h]
    \centering
    \includegraphics[width=\linewidth]{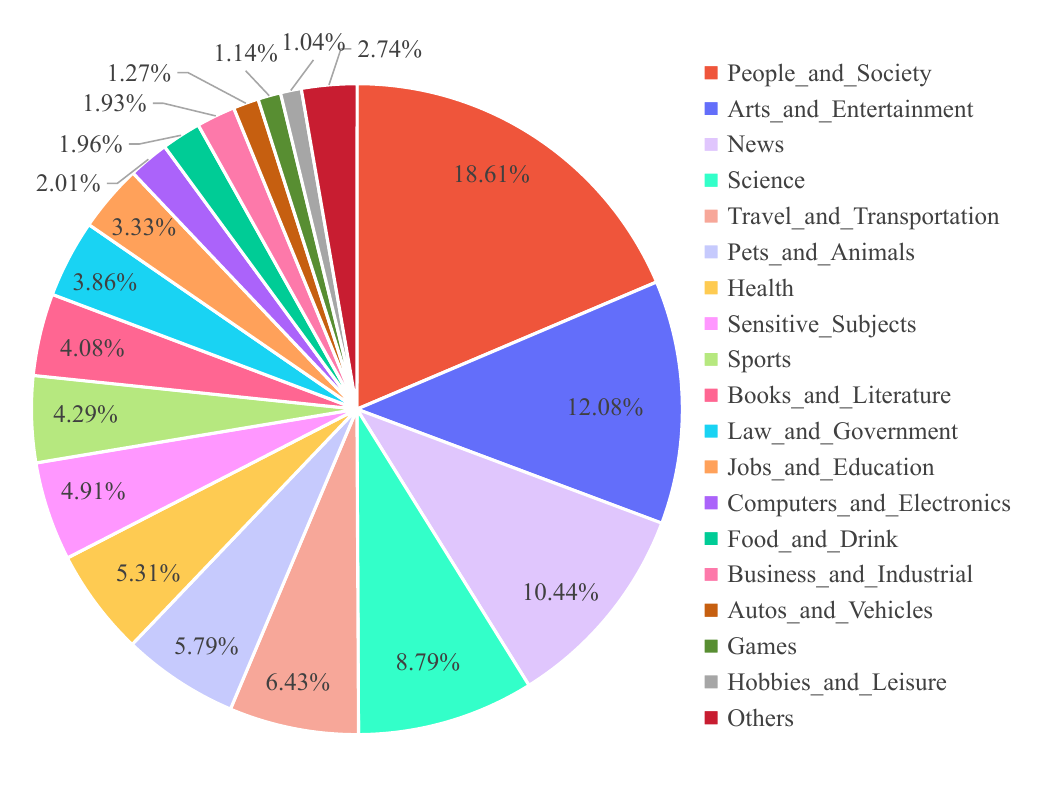}
    \caption{Domain Distribution in \textsc{MTR-Bench}. The dataset covers a diverse range of subjects, rigorous testing generalization.}
    \label{fig:topic}
\end{figure}

\section{\textsc{MTR-Bench}: A Realistic Conversational Benchmark}
\label{sec:mtr_bench}

Leveraging the \textsc{MTR-Pipeline}, we synthesized \textsc{MTR-Bench}, a large-scale, general-domain conversational retrieval benchmark. Unlike previous datasets restricted by human cognitive bottlenecks or rigid heuristic rules, \textsc{MTR-Bench} is explicitly engineered to stress-test modern RAG systems against the complexities of real-world production environments. In this section, we analyze the statistical characteristics of the benchmark and detail the design principles that contribute to its high discriminative difficulty. Table \ref{tab:database_info} lists the key parameters of \textsc{MTR-Bench} and compares them with previous benchmarks. In the following sections, we will elaborate on how these parameters influence the relevance of the benchmark to real-world RAG scenarios. We also provide engineering guidelines to assist users in selecting these parameters for their specific scenarios, thereby enhancing the effectiveness of evaluations using our pipeline (see Appendix \ref{appendix:MTR-BENCH}).

\subsection{Dataset Overview and Statistics}
\label{subsec:dataset_statistics}

\textsc{MTR-Bench} is constructed from the Wikipedia 2025-01 dump, ensuring temporal relevance. The dataset is split into a training set (MTR-train) and an evaluation benchmark  (\textsc{MTR-Bench}). As shown in Table~\ref{tab:dataset-analysis}, the benchmark comprises 1,000 conversations with 8,000 turns, featuring an average of 5.6 distinct topics per conversation, which indicates high informational density.

\paragraph{Domain Diversity}
To ensure robust evaluation, a general-purpose benchmark must cover a wide spectrum of knowledge. We analyzed the semantic distribution of \textsc{MTR-Bench} using a domain classifier. As shown in Figure~\ref{fig:topic}, the dataset achieves broad coverage across Humanities, STEM, Social Sciences, and Vital Statistics. This diversity is critical for testing a retriever's ability to generalize across distinct contexts, rather than overfitting to specific domains.

\paragraph{Topic Flow Dynamics}
A key feature of \textsc{MTR-Bench} is its complex topic evolution. We visualize the inter-turn topical transitions using a Sankey diagram in Figure~\ref{fig:topic_flow}. The visualization reveals that conversations do not merely drift linearly; they exhibit realistic recursive patterns where users revert to previously discussed subjects or branch into distinct sub-topics. This non-linear flow challenges a retriever's ability to maintain long-range context dependence.

\subsection{Why is \textsc{MTR-Bench} Hard?}
\label{subsec:design_principles}
We explicitly engineered three factors that increase difficulty:

\subsubsection{Realistic Topic Switching (Hard vs. Soft)}
Human information-seeking is rarely linear. Research indicates that users switch topics approximately every 2.11 turns \citep{Spink2002MultitaskingIS}. Most existing benchmarks capture only smooth transitions. \textsc{MTR-Bench} implements two distinct switching modes:

\paragraph{Soft Topic Switching:} Simulated via Wikipedia hyperlinks, these transitions represent natural exploratory behavior where a user navigates between semantically related concepts.
\paragraph{Hard Topic Switching:} This mode simulates abrupt context shifts, akin to a user initiating a completely new query without clearing the chat history. This creates severe contextual interference, forcing the retriever to distinguish between the current query's intent and irrelevant global history.

\subsubsection{Production-Grade Response Characteristics}
\label{subsubsec:response_style}
A major discrepancy exists between academic datasets (often annotated by humans who prefer brevity) and production environments (dominated by verbose LLMs).

\paragraph{Long-Context Responses}
LLMs driven by RLHF \citep{ouyang2022traininglanguagemodelsfollow}, tend to be verbose. Analysis of real-world ShareGPT data reveals an average assistant response length of $\sim$464 tokens, whereas traditional benchmarks average far fewer. \textsc{MTR-Bench} mirrors this reality (see Table \ref{tab:dataset-analysis}), forcing retrievers to process substantially more noise within the dialogue history.

\paragraph{Ambiguous Decision Style}
Traditional datasets often handle ambiguity through clarifying questions. This approach lowers retrieval difficulty because the subsequent user turn is typically a simple confirmation (e.g., yes or no), allowing the system to retain the previous gold document. In contrast, \textsc{MTR-Bench} mimics production LLMs, which tend to proactively guess the user's intent to provide an immediate answer. This eliminates low-entropy clarification turns and instead fills the history with verbose, speculative content, forcing the retriever to filter through dense noise rather than relying on a static context.

\subsubsection{Industrial-Scale Knowledge}
\paragraph{Knowledge Cutoff (2025.01)}
Pre-trained models often rely on internal parametric memory for older facts. To strictly evaluate retrieval capability rather than memorization, \textsc{MTR-Bench} utilizes a very recent Wikipedia dump (2025.01). This targets ``long-tail'' and recent knowledge \citep{kandpal2023largelanguagemodelsstruggle} that models have not seen during training.

\paragraph{Optimized Document Granularity}
We set a target document length of $\sim$1024 characters. This choice is informed by LlamaIndex research, balancing the need for sufficient context (avoiding fragmentation) with the limits of embedding model fidelity \citep{zhu-etal-2024-longembed}. This granularity ensures that the challenge stems from semantic matching rather than arbitrary chunking artifacts.

\section{Experiments}
\subsection{Models}

We selected a diverse set of dense retrievers for evaluation,
the BERT-sized SOTA shared encoder \texttt{stella\_en\_400m\_v5}~\citep{zhang2025jasperstelladistillationsota},
the community-popular \texttt{bge-large-en-v1.5}~\citep{bge_embedding},
and \texttt{gte-modernbert-base}, which has demonstrated strong performance on the novel ModernBERT~\citep{modernbert} architecture.
In addition to these single-turn retrievers, we incorporate existing SOTA Conversational Dense Retrievers (CDRs) specifically designed to handle multi-turn dialogue context, including \texttt{Dragon-DocChat}\footnote{\url{https://huggingface.co/cerebras/Dragon-DocChat-Context-Encoder}} and \texttt{Dragon-ChatQA}~\citep{liu2024chatqa}.
These models were trained on a synthetic dataset derived from \texttt{ChatQA}~\citep{liu2024chatqa} on \texttt{Dragon+}~\citep{lin2023traindragondiverseaugmentation} with different implementations.

\subsection{Metrics}
We employed the commonly used retrieval performance evaluation metrics, Recall (@5, @20), NDCG@20, and MRR@20. Recall only considers whether the golden document is included within a given retrieval budget (@k) but is highly interpretable. NDCG and MRR build upon this by also being sensitive to the golden document's position in the retrieved set, offering a more fine-grained representation. We provide the NDCG and MRR scores in Table~\ref{tab:mrr_ndcg}.

\begin{table*}[h]

\resizebox{\textwidth}{!}{  
\begingroup
  \setlength{\tabcolsep}{3pt}
  \tiny
  \centering
  \tiny
  \begin{tabular}{l|
                  c|c| c|c| c|
                  c|c| c|c| c|c|c}
    \toprule
    & \multicolumn{10}{c}{\textbf{SOTA Single-turn Retriever}} \\
    \cmidrule{2-13}
    \multicolumn{1}{l|}{\multirow{1}{*}{\textbf{Model} }}
      & \multicolumn{2}{c|}{\textbf{QReCC}}
      & \multicolumn{2}{c|}{\textbf{QuAC}}
      & \multicolumn{2}{c|}{\textbf{Doc2Dial}}
      & \multicolumn{2}{c|}{\textbf{TopiOCQA}} 
      & \multicolumn{2}{c|}{\textbf{MTR-BENCH}} 
      & \multicolumn{2}{c}{\textbf{Average}} 
      \\
    \cmidrule(lr){2-3} \cmidrule(lr){4-5} 
    \cmidrule(lr){6-7} \cmidrule(lr){8-9} \cmidrule(lr){10-11} \cmidrule(lr){12-13} 
    
& \textbf{R@5} & \textbf{R@20} & \textbf{R@5} & \textbf{R@20} &\textbf{R@5} & \textbf{R@20} &\textbf{R@5} & \textbf{R@20} &\textbf{R@5} & \textbf{R@20}&\textbf{R@5} & \textbf{R@20}  \\
    \midrule

    \texttt{gte-Qwen2-7B}
      & 59.78  & 90.58 
  & 73.74 
  & 94.97 
  &  80.02 
  &  93.58 
       &    69.29 
     &  89.46 
       &     39.75    &  53.23 &64.52 &	84.36 
       \\
    \texttt{stella\_en\_400m\_v5}
      & 92.41  & 99.21  & 81.13  & 94.41 &81.54   &  94.16  &  45.27  &    68.58 &      39.38 &      47.30 &67.94 &	80.73  \\
   \texttt{bge-large-en-v1.5}
      & 80.23  & 96.74  & 74.33   & 92.02   & 66.21  
      & 87.05    & 42.44 
        &   64.08 
      &   30.16 
      &   35.04 &58.67 &	74.99
      \\
    \texttt{gte-modernbert-base}
      & 92.44 
  & 99.10 
  & 83.74 
  & 95.95 
  & 80.02 

      & 93.58 
        & 71.04 
        & 89.70 
        & 50.29 
        & 59.31  &75.51 &	87.53 
        \\
    \midrule
    & \multicolumn{10}{c}{\textbf{SOTA Conversational Dense Retriever}} \\
    \cmidrule{2-13}
    \multicolumn{1}{l|}{\multirow{1}{*}{\textbf{Model}}}
      & \multicolumn{2}{c|}{\textbf{QReCC}}
      & \multicolumn{2}{c|}{\textbf{QuAC}}
      & \multicolumn{2}{c|}{\textbf{Doc2Dial}}
      & \multicolumn{2}{c|}{\textbf{TopiOCQA}} 
      & \multicolumn{2}{c|}{\textbf{MTR-BENCH}} 
      & \multicolumn{2}{c}{\textbf{Average}}  \\
    \cmidrule(lr){2-3} \cmidrule(lr){4-5} 
    \cmidrule(lr){6-7} \cmidrule(lr){8-9} \cmidrule(lr){10-11} 
    \cmidrule(lr){12-13} 
 & \textbf{R@5} & \textbf{R@20} & \textbf{R@5} & \textbf{R@20} &\textbf{R@5} & \textbf{R@20} &\textbf{R@5} & \textbf{R@20} &\textbf{R@5} & \textbf{R@20} &\textbf{R@5} & \textbf{R@20} \\
    \midrule

    \texttt{Dragon-ChatQA}
      & 91.76  & 98.96  & 86.13  & 96.55   & 83.50 & 95.40  & 66.35  &  84.81  & 43.84  & 50.96     &74.32 &	85.34        \\
    \texttt{Dragon-DocChat}
      & 91.51 
 & 99.03 
& 80.55 &	95.24 &	78.62 &	93.15 	&71.96 	&85.80 &	31.50 &	40.98 &70.83 &	82.84 
   \\
      

    \bottomrule
  \end{tabular}
  \label{tab:results}
\endgroup

}
\caption{Recall (R@k) performance of various retrievers across multiple conversational retrieval benchmarks. We use R as a general shorthand for Recall.}
\label{tab:main_result}
\end{table*}

\subsection{Results}
For each evaluation instance in the benchmark, the input context is constructed by serializing the conversation history and prepending it to the current query. The history is formatted by alternating ``\texttt{User:}'' and ``\texttt{Agent:}'' prefixes for each turn. We present our primary experimental findings in Table \ref{tab:main_result}. The results lead to the following key observations:

\paragraph{\textsc{MTR-Bench} offers substantial headroom for retriever improvement.}
    A consistent observation is that \textsc{MTR-Bench} presents a more challenging evaluation landscape compared to many existing conversational retrieval benchmarks. While leading models achieve Recall@20 scores exceeding 90 points on several prior datasets, their average Recall@20 on these benchmarks is 43.54 points higher than their average performance on \textsc{MTR-Bench}. This significant performance differential suggests that \textsc{MTR-Bench} is less susceptible to score saturation and provides greater capacity to differentiate the capabilities of current and future advanced retrieval systems.
    
    \paragraph{\textsc{MTR-Bench} probes deeper retrieval capabilities.}
    The benchmark's challenging nature is further validated by the marginal performance gains when increasing the recall budget. On average, expanding the retrieval window from the top 5 to the top 20 documents (Recall@5 vs. Recall@20) yields a 15.06 point improvement in Recall@20 on previous benchmarks. In contrast, the same expansion on \textsc{MTR-Bench} results in a more modest average improvement of only 8.68 points. This suggests that \textsc{MTR-Bench} demands more precise and robust retrieval, as simple increases in the number of retrieved candidates provide diminishing returns, indicating a more complex relevance landscape.

\section{Analysis}

\begin{table}[t]
\centering
\setlength{\tabcolsep}{3.5pt} 

\resizebox{\linewidth}{!}{ 
    \begin{tabular}{lcccc|cc}
    \toprule
    & \multicolumn{4}{c|}{\textsc{MTR-EVAL Metrics}} & \multicolumn{2}{c}{\textbf{Retrieval (BGE)}} \\
    \textbf{Dataset Setting} & \textbf{Comp.} & \textbf{Q-E} & \textbf{A-E} & \textbf{Qual.} & \textbf{R@5} & \textbf{R@20} \\
    \midrule
    \textbf{MTR-FINANCE} & 4.50 & 4.54 & 4.70 & 4.91 & 0.37 & 0.50 \\
    \textit{w/o Filter} & 4.67 & 4.72 & 4.82 & 4.90 & 0.45 & 0.56 \\
    \bottomrule
    \end{tabular}
}
\caption{Validation on the industrial-scale MTR-FINANCE dataset. Top row: The full pipeline generates high-quality benchmarks that remain challenging for retrievers (low Recall). Bottom row: Removing the quality filter results in higher recall and evaluation scores, indicating the inclusion of simpler, less discriminative documents.}
\label{tab:finance_ablation}
\end{table}

\subsection{Robustness Across Domains}
A critical requirement for automated benchmark synthesis is transferability to domain-specific or proprietary knowledge bases. To validate this, we applied \textsc{MTR-Pipeline} to an internal industrial-scale financial corpus. Crucially, this underlying knowledge base is highly complex and heterogeneous, comprising a dense mixture of unstructured emails, formal regulations, legal charters, and granular transaction records. Due to the proprietary nature of the source data, the raw corpus remains private; however, we plan to partially release the generated dialogues to serve as a reference in future work.

As shown in the first row of Table~\ref{tab:finance_ablation}, the synthetic financial benchmark maintains high annotation quality, with an Answer Quality score of 4.91 and Answer-Evidence Faithfulness of 4.70. Conversely, the retrieval difficulty remains substantial; the BGE retriever achieves a Recall@5 of only 0.37 on \textsc{MTR-Bench-finance}, compared to much higher scores on simpler datasets. This low retrieval score is attributable to the intricate nature of the mixed data sources. This demonstrates that \textsc{MTR-Suite} successfully generates challenging, high-quality evaluation instances for specialized domains, confirming its viability for auditing enterprise RAG systems beyond general-domain Wikipedia data.

\subsection{Ablation Study}
We performed an ablation study to quantify the contributions of the \textit{Filtering} and \textit{Polisher} modules, using \textsc{MTR-Bench-finance} as the testbed.

\paragraph{Impact of Filtering.} We compared the full pipeline against a version where the quality filter was removed. As shown in Table~\ref{tab:finance_ablation} (\textit{w/o Filter}), removing the filter resulted in increased retrieval recall (R@5 rose from 0.37 to 0.45) and slightly higher automated evaluation scores. While higher scores might initially seem positive, qualitative analysis reveals that the unfiltered corpus contains simpler, less information-dense passages that are trivial to retrieve and question. The filtering module is therefore essential for enforcing a high difficulty standard and ensuring the benchmark targets complex, educational content.

\paragraph{Impact of Polisher.} To evaluate the Polisher, we utilized the human evaluation setup described in Section 7.2. When the Polisher module was removed from the pipeline, the accuracy of human annotators in identifying machine-generated questions increased significantly from 62\% to 79\%. This indicates that the Polisher plays a crucial role in linguistic refinement, smoothing out the structural rigidity of the raw Questioner agent to produce natural, human-like dialogue.

\subsection{Disentangling Knowledge Base Recency from Dialogue Complexity}
\label{subsec:disentangle}

Since \textsc{MTR-Bench} uses a recent Wikipedia dump (2025.01) unseen during retriever pre-training, a natural concern is whether the observed performance degradation stems primarily from unfamiliar knowledge or from the linguistic complexity of conversational queries. We provide two pieces of evidence to disentangle these factors.

\paragraph{Oracle Query Rewriting.} Our LLM-based query rewriting analysis (Figure~\ref{fig:rewrite}) reveals a 20\%--40\% R@5 improvement when raw conversational queries are rewritten into explicit, self-contained forms. Crucially, the rewritten queries achieve high absolute recall \textit{on the same 2025.01 knowledge base}, demonstrating that gold documents are readily retrievable when query intent is made explicit. This confirms that the difficulty is not primarily caused by the knowledge base being unfamiliar to the retrievers, but rather by the linguistic complexity of the raw conversational queries, ellipsis, coreference, topic switching, and verbose history.

\paragraph{Cross-Domain Consistency.} When \textsc{MTR-Pipeline} is applied to a completely different corpus (the financial domain in Section~\ref{subsec:dataset_statistics}), retrievers exhibit similarly low performance despite the corpus having its own temporal characteristics unrelated to Wikipedia recency. This further supports that the pipeline's dialogue design is the primary driver of retrieval difficulty.

Together, these results confirm that \textsc{MTR-Bench}'s challenge arises predominantly from genuine conversational complexity rather than knowledge base artifacts.

\subsection{Extended Recall@k Analysis}
\label{appendix:extended_recall}

To assess retrieval performance under large candidate budgets---relevant for RAG pipelines using long-context reader models that can consume hundreds or thousands of documents---we compute extended Recall@k scores on \textsc{MTR-Bench}. Results are presented in Table~\ref{tab:extended_recall}.

\begin{table}[h]
\centering
\small
\setlength{\tabcolsep}{3.5pt}
\resizebox{\linewidth}{!}{
\begin{tabular}{lcccccc}
\toprule
\textbf{Model} & \textbf{R@1} & \textbf{R@5} & \textbf{R@20} & \textbf{R@100} & \textbf{R@500} & \textbf{R@1000} \\
\midrule
bge-large-en-v1.5 & 20.2 & 30.2 & 35.0 & 40.0 & 45.0 & 47.0 \\
ChatQA-Context & 29.1 & 43.8 & 51.0 & 58.0 & 64.7 & 67.4 \\
gte-Qwen2-7B-instruct & 21.1 & 39.8 & 53.2 & 66.5 & 78.2 & 82.2 \\
\bottomrule
\end{tabular}
}
\caption{Extended Recall@k on \textsc{MTR-Bench}. Even at $k{=}1000$, no model achieves full recall, confirming that a substantial fraction of gold documents are entirely outside the retriever's representational reach.}
\label{tab:extended_recall}
\end{table}

Several observations emerge. First, \textsc{MTR-Bench} remains challenging even at very large $k$: the best model (gte-Qwen2-7B) reaches only 82.2\% at R@1000, while bge-large saturates at 47.0\%. Second, model-dependent scaling behavior is evident: the 7B-parameter model benefits substantially from increasing $k$ (R@20$\to$R@1000: $+$29.0 points), suggesting it places relevant documents in the broader candidate set but struggles with precise ranking. In contrast, bge-large shows diminishing returns ($+$12.0 points), indicating many gold documents are entirely outside its representational capacity. Third, for RAG pipelines using a long-context reader over the top 100 retrieved documents, even the best retriever still misses $\sim$34\% of relevant evidence, a hard ceiling that no amount of reader sophistication can overcome.

\section{Conclusion}

This paper presents \textsc{MTR-Suite}, a scalable framework for high-fidelity conversational retrieval benchmarking. By anchoring \textsc{MTR-Eval} to human-validated metrics and constraining \textsc{MTR-Pipeline} with greedy traversal clustering, we decouple data complexity from model-specific biases.  Crucially, in enterprise environments where knowledge bases evolve rapidly, manual annotation often serves as a lagging indicator. \textsc{MTR-Suite} addresses this by enabling continuous, on-demand benchmarking synchronized with data updates. Our experiments show that while \textsc{MTR-Eval} confirms the linguistic quality of our data, the significant performance drop in state-of-the-art retrievers proves that \textsc{MTR-Bench} introduces genuine semantic challenges rather than artifacts of LLM favoritism. \textsc{MTR-Suite} thus provides a cost-effective, rigorous standard for evaluating RAG systems without the risk of circular reasoning.

\section{Acknowledgement}
This work was supported in part by the National Natural Science Foundation of China (Nos. U24A20334 and 62276056), the Yunnan Fundamental Research Projects (No. 202401BC070021), the Yunnan Science and Technology Major Project (No. 202502AD080014), the Fundamental Research Funds for the Central Universities (Nos. N25BSS054 and N25BSS094), and the Program of Introducing Talents of Discipline to Universities, Plan 111 (No. B16009).

\clearpage

\section{Limitations}
\label{sec:limitations}
\textsc{MTR-Bench} focuses explicitly on the retrieval component rather than end-to-end (E2E) generation. This is a deliberate choice to preserve conversational realism: standard E2E metrics (e.g., Exact Match, BLEU) require rigid formatting or short ground-truth answers, which contradict the verbose and explanatory nature of real-world RAG responses. By avoiding these artificial constraints, we ensure the benchmark reflects actual application utility. Given that retrieval quality dictates the factual upper bound of RAG systems, we prioritize a precise diagnostic of the retriever over a potentially noisy and constrained E2E evaluation.
\section{Ethical Statement}

\label{sec:ethics}

As we release \textsc{MTR-Bench} to the research community, ensuring the safety, compliance, and ethical integrity of the dataset is paramount. The harmlessness of the generated data is guaranteed through a three-tiered framework:

\paragraph{Source Data Hygiene:}

Our document corpus is derived exclusively from Wikipedia. While Wikipedia inherently adheres to strict community guidelines that prohibit explicit sexual or violent content, we implement an additional layer of safety during preprocessing. As detailed in our pipeline description, all documents pass through a rigorous educational quality filter. This step automatically identifies and removes any residual entries containing sensitive, offensive, or non-educational material, ensuring that the underlying knowledge base remains strictly knowledge-centric and ``clean'' before it enters the generation phase.

\paragraph{Model Alignment:}

The synthetic data generation is powered by state-of-the-art Large Language Models (LLMs) that have undergone extensive safety alignment (e.g., RLHF) to reject harmful instructions. Furthermore, our prompt engineering is meticulously designed to be task-specific. By focusing the agents strictly on information retrieval and reasoning tasks, we minimize the risk of "jailbreaking" the models or inducing the generation of toxic, biased, or inappropriate content.

\paragraph{Constrained Generation:}

Unlike open-ended chitchat systems, the generation process in \textsc{MTR-Pipeline} is not free-form. The interactions are strictly grounded in the prstyovided reference documents. The agents are instructed to formulate questions and answers solely based on the filtered evidence present in the corpus. This constraint acts as a final safeguard, preventing the models from hallucinating harmful content or introducing external biases unrelated to the source text.

Collectively, these measures ensure that \textsc{MTR-Bench} is compliant with ethical standards and safe for broad academic use.

\bibliography{latex/acl}
\appendix

\clearpage
\section{Appendix}
\label{sec:appendix}

\subsection{Previous Benchmark Evaluation}
\label{app:previous_benchmark_heatmap}
\begin{figure*}[!htbp] 
    \centering
    
    
    \begin{subfigure}{0.48\textwidth}
        \centering
        \includegraphics[width=\linewidth]{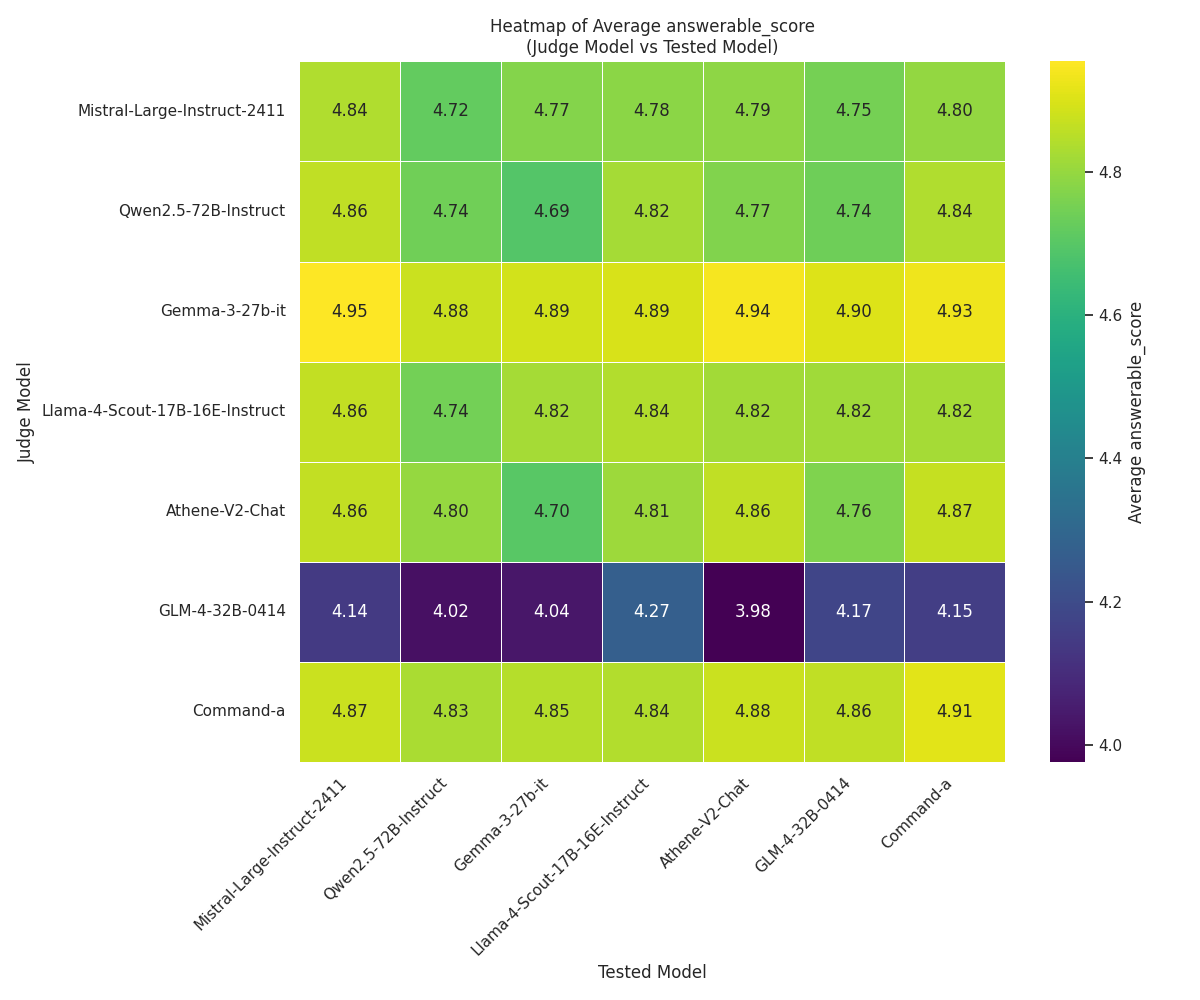}
        \caption{Query relevance heatmap visualization.}
        \label{fig:answerable_score}
    \end{subfigure}
    \hfill 
    \begin{subfigure}{0.48\textwidth}
        \centering
        \includegraphics[width=\linewidth]{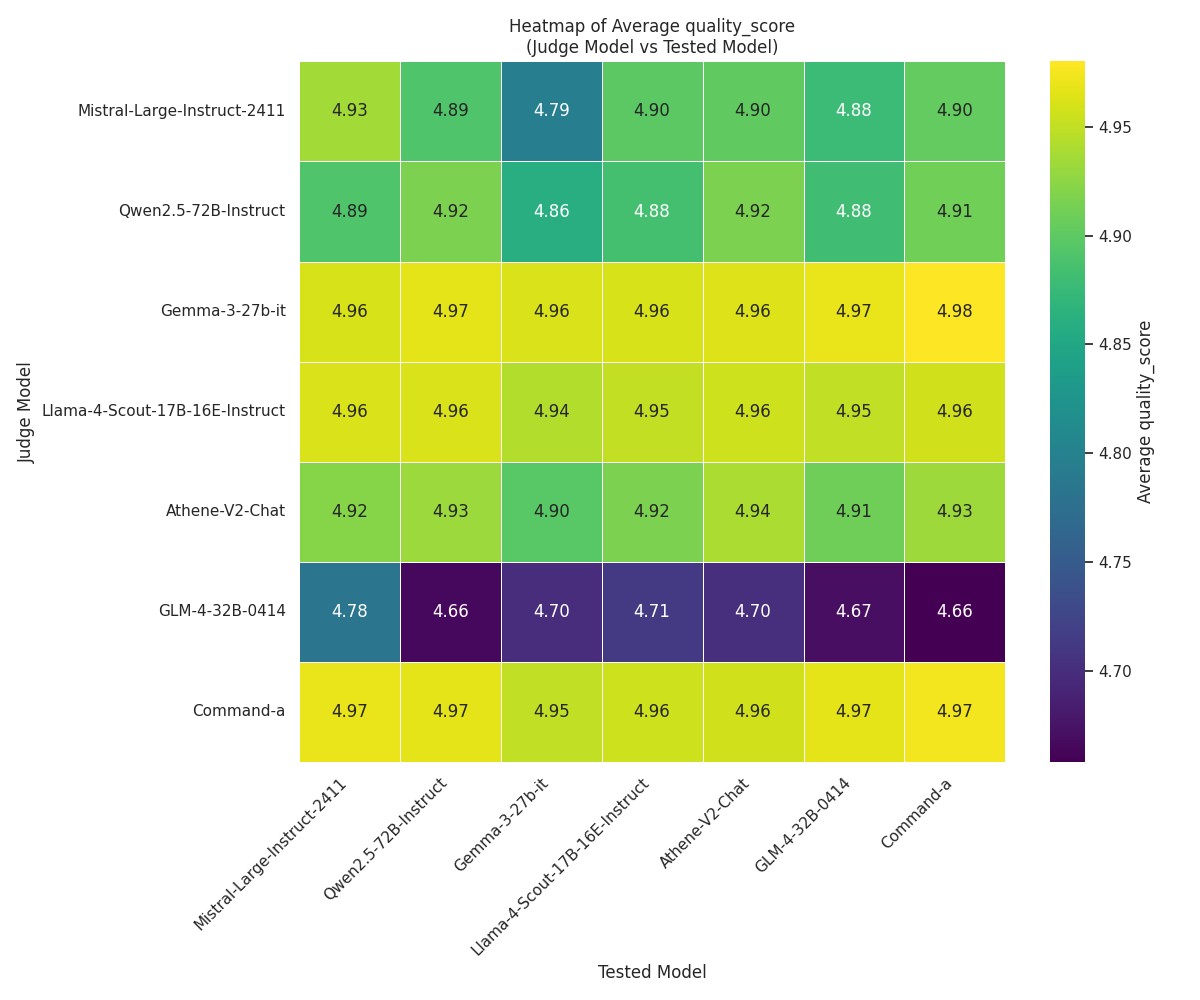}
        \caption{Quality Score heatmap visualization.}
        \label{fig:quality_score}
    \end{subfigure}
    
    \vspace{1em} 
    
    
    \begin{subfigure}{0.48\textwidth}
        \centering
        \includegraphics[width=\linewidth]{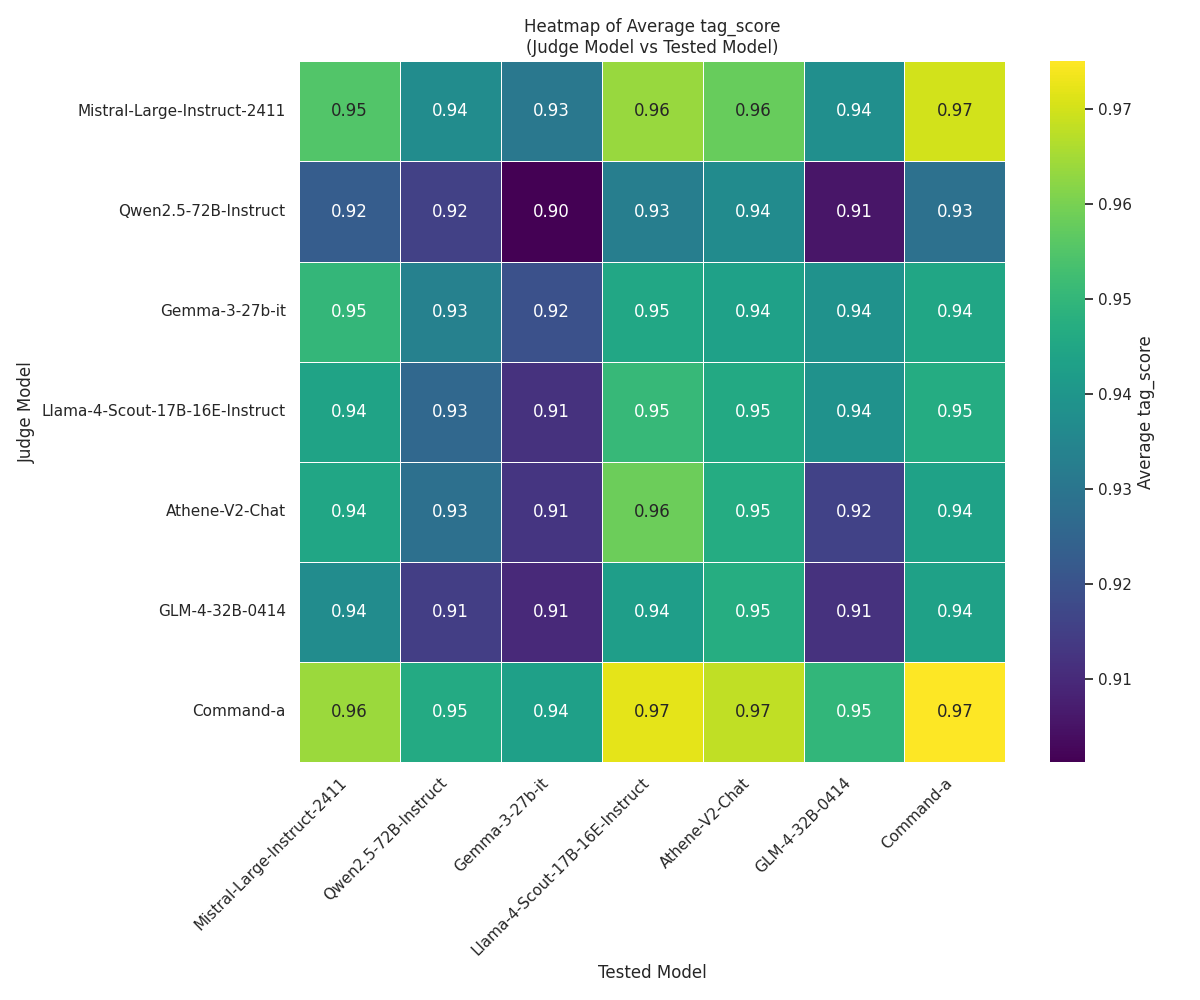}
        \caption{Annotation accuracy heatmap visualization.}
        \label{fig:tag_score}
    \end{subfigure}
    \hfill
    \begin{subfigure}{0.48\textwidth}
        \centering
        \includegraphics[width=\linewidth]{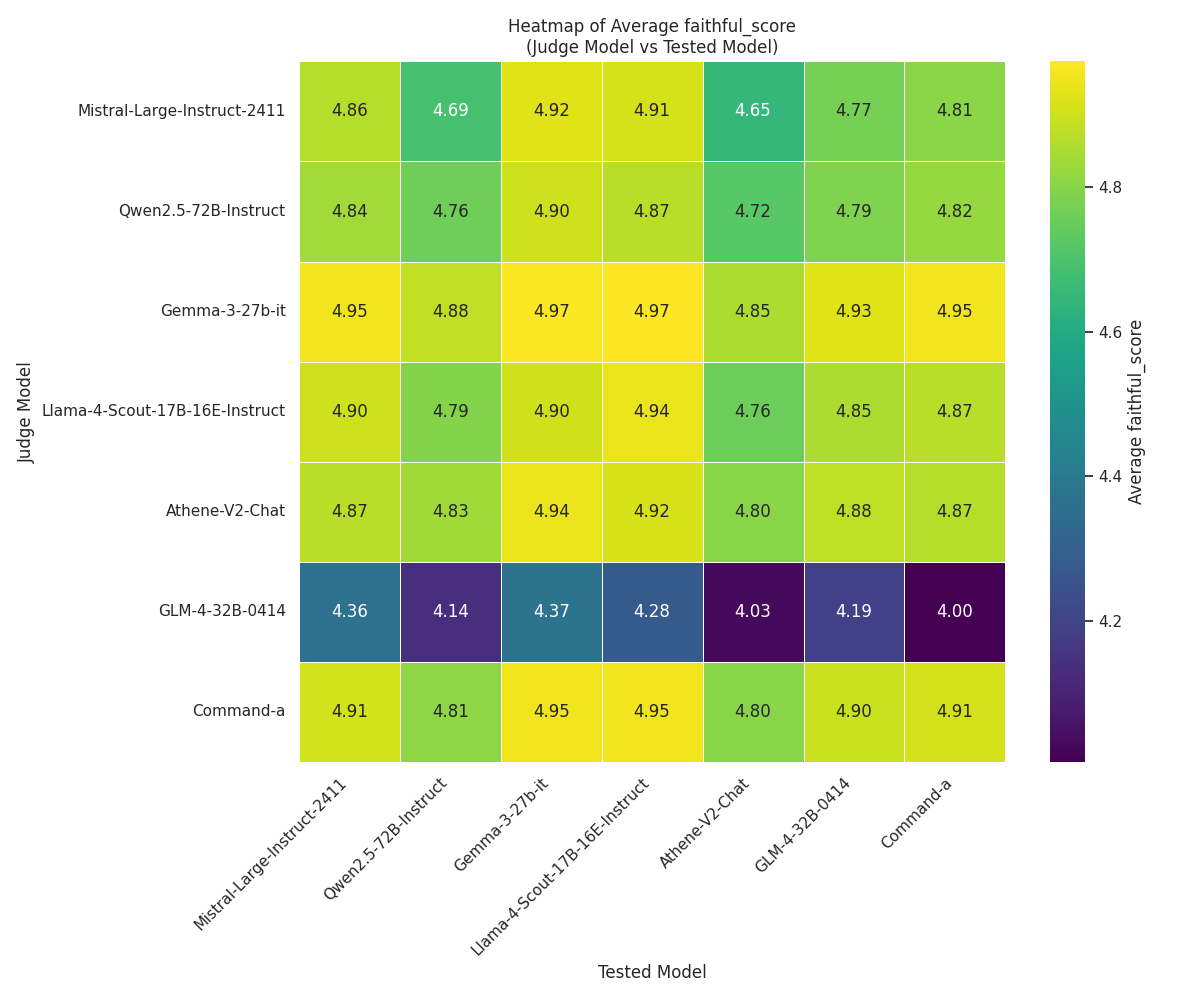}
        \caption{Response Faithful heatmap visualization.}
        \label{fig:faithful_score}
    \end{subfigure}
    
    \caption{Summary of MTR-eval scores for different models on previous benchmarks.}
    \label{fig:all_heatmaps}
\end{figure*}

For each historical benchmark considered in our evaluation, we elicited four distinct scores from each of the seven selected models. This systematic process yielded four 7x6 model-benchmark heatmaps, providing a granular view of relative performance across different tasks, as depicted in \cref{fig:answerable_score,fig:quality_score,fig:faithful_score,fig:tag_score}.

\subsection{Quality Audit and Limitations of the CORAL Benchmark} 
\label{appendix: coral}

CORAL constructs a hierarchical title tree by leveraging the structure of Wikipedia, such as titles and subtitles. By traversing this tree, it generates answers at various levels. The method utilizes GPT-4 to rewrite titles into user questions, while answers are extracted directly from the body text under the corresponding titles. Gold documents are identified from the references cited within the text. This pipeline relies heavily on documents possessing a clear hierarchical structure, which constrains its utility as a general-purpose automated process. Additionally, the approach for constructing the (question, answer, support document) tuples is relatively simplistic, resulting in notable quality concerns.

Our preliminary evaluation indicated low performance on the CORAL benchmark. A detailed case study revealed that the dataset exhibits notable annotation inconsistencies, which may affect its reliability for fine-grained benchmarking. The first data sample from the dataset serves as an illustration:

\begin{quote}
    \textbf{U:} What methods are used to predict these geodynamic regimes? \\
    \textbf{A:} exoplanets have been directly observed and remotely sensed ,[15] but due to their great distance and proximity to obscuring energy sources (the stars they orbit), there is little concrete knowledge of their composition and geodynamic regime. therefore, the majority of information and conjectures made about them come from alternative sources \\
    \textbf{support document:} \{'document\_idx': 15, 'ctx': ``doi it uses a managed registry ( providing social and technical infrastructure ) . it does not assume any specific business model for the provision of identifiers or services and enables other existing services to link to it in defined ways . several approaches for making identifiers persistent have been proposed . the comparison of persistent identifier approaches is difficult because they are not all doing the same thing . imprecisely referring to a set of schemes as `` identifiers '' doesn't mean that they can be compared easily . other `` identifier systems '' may be enabling technologies with low barriers to entry , providing an easy to use labeling mechanism that allows anyone to set up a new instance ( examples include persistent uniform resource locator , urls , globally unique identifiers , etc . ) , but may lack some of the functionality of a registry-controlled scheme and will usually lack accompanying metadata in a controlled scheme . the doi system does not have this approach and should not be compared directly to such identifier schemes . various applications using such enabling technologies with added features have been devised that meet some of the features offered by the doi system for specific sectors ( e.g. , ark ) . a doi name does not depend on the object 's location and , in this way , is similar to a uniform resource name or purl but differs from an ordinary url ."\}
\end{quote}

The dialogue pertains to astronomy, while the support document discusses the Digital Object Identifier (DOI) system. This misalignment illustrates the type of query-evidence inconsistency that can arise from rule-based synthesis pipelines relying on document structure.

We communicated these findings to the CORAL team. Following our discussion and the provision of quantitative and qualitative evidence, Following constructive discussion, the authors proactively revised their dataset and released an improved version (CORAL v2), demonstrating the practical utility of \textsc{MTR-Eval} as a diagnostic tool for benchmark quality assurance. The authors later released a revised version (hereafter CORAL v2) and requested further testing. However, since CORAL v2 is not the dataset associated with the original publication, we present results based on the original version. It is worth noting that while performance on CORAL v2 improved, the scores remained lower than those achieved on preceding manually annotated datasets.

This discussion highlights two primary conclusions:
\begin{enumerate}
\item Fully automated benchmark synthesis for conversational retrieval remains a challenging open problem, and quality assurance mechanisms are essential.
\item \textsc{MTR-Eval} can serve as an effective and actionable diagnostic tool, facilitating iterative improvement of benchmark quality across research groups.
\end{enumerate}

\subsection{Case Study on Annotation Cognition Boundaries}
\label{app:case_study}

Human annotation is inherently limited by a "Cognition Boundary." Annotators typically formulate queries based on the specific document they are reading at that moment. However, in a large-scale corpus, the "Gold" document selected by the annotator is often not the unique or exclusive source of the answer. This results in \textit{Annotation Sparsity}, where valid supporting documents are overlooked simply because the annotator did not see them.

To illustrate this, Table \ref{tab:case_study} compares the human-annotated "Gold" documents with \textbf{alternative valid documents} identified within the corpus. These examples clearly show that ground truth is rarely unique:

\begin{itemize}
    \item \textbf{Same Answer in Different Documents:} In datasets like \textbf{TopiOCQA} and \textbf{InSCIT}, the information is not unique to the Gold document. As shown in the table, the alternative documents retrieved by the model contain the exact same answer (e.g., the date "1980" or the country "Brazil") as the human-annotated text. Both sources are equally correct.

    \item \textbf{More Direct or Suitable Answers:} In other cases, such as \textbf{QReCC}, \textbf{Doc2Dial}, and \textbf{QuAC}, the alternative document found in the corpus may actually address the user's specific question more directly. For instance, in \textbf{QReCC}, the user explicitly asks for "types" of heartbeats. The human annotator selected a document discussing \textit{causes}, whereas the alternative document provides a specific list of \textit{types}. Similarly, in \textbf{Doc2Dial}, the alternative document defines the "Retirement Planner" itself, rather than the "Online Calculator" selected by the annotator.
\end{itemize}

These cases demonstrate that the human-annotated "Gold" document is often just \textit{one of several} valid options, and sometimes not even the most direct one. Therefore, strict evaluation metrics that penalize models for retrieving these valid alternatives may underreport a system's true capabilities.

\begin{table*}[ht]
\small
\centering
\renewcommand{\arraystretch}{1.5}
\begin{tabularx}{\textwidth}{p{1.6cm} p{3.2cm} X X}
\toprule
\textbf{Dataset} & \textbf{Query} & \textbf{Annotated Evidence (Gold)} & \textbf{Alternative Valid Evidence (Retrieved)} \\
\midrule

\textbf{QReCC} & 
\textit{Tell me about the types of irregular heart beat.} & 
\dots Rhythm disturbances may be normal physiologic responses \dots Lack of oxygen can occur when the lungs are unable to extract oxygen \dots Significant anemia \dots decreases the oxygen-carrying capacity \dots & 
\dots Heart rhythm disorders are classified according to where they occur \dots \textbf{There are twelve types of heart rhythm disorders} \dots Atrial fibrillation occurs when the atrium has lost the ability to beat in a coordinated fashion \dots \\
\midrule

\textbf{Doc2Dial} & 
\textit{I want to get some info about the retirement benefits planner.} & 
Benefits Planner: Retirement \textbf{Online Calculator} (WEP Version). The calculator shown below allows you to estimate your Social Security benefit \dots Note: If your birthday is on January 1st \dots & 
\textbf{Benefits Planner} \dots Use our planners to help you better understand your Social Security protection \dots \textbf{Retirement Benefits: Use our Retirement Planner to learn}: how you qualify \dots about possible benefits \dots \\
\midrule

\textbf{TopiOCQA} & 
\textit{when did victor start on young and restless} & 
William J. Bell created Victor as a short-term non-contractual role, \textbf{debuting on February 8, 1980}. Bell stated in 1997 \dots & 
 Victor Newman is a fictional character \dots He has been \textbf{portrayed by Eric Braeden since 1980}. Initially a guest character who was to last for eight to twelve weeks \dots \\
\midrule

\textbf{InSCIT} & 
\textit{What is the only country that has to disclose trace amounts of allergens \dots?} & 
\dots Nevertheless, there are no labeling laws to mandatory declare the presence of trace amounts \dots \textbf{except in Brazil}. & 
 \textbf{In Brazil since April 2016}, the declaration of the possibility of cross-contamination is mandatory when the product does not intentionally add any allergenic food \dots \\
\midrule

\textbf{QuAC} & 
\textit{Was death of a Ladies man an album?} & 
Spector began to reemerge \dots producing and co-writing a controversial 1977 \textbf{album by Leonard Cohen, entitled Death of a Ladies' Man}. This angered many devout Cohen fans \dots & 
 All songs written by Leonard Cohen \dots Categories: \textbf{1977 albums}, Leonard Cohen albums \dots Cohen published the book Death of a Lady's Man in 1978. It has nothing in common with the album \dots \\

\bottomrule
\end{tabularx}
\caption{Case study illustrating annotation sparsity. In each instance, we identify an \textbf{Alternative Valid Document} from the corpus that differs from the human-annotated Gold standard. These alternatives often answer the query more directly or provide complementary details, highlighting that relevant evidence in large corpora is rarely unique.}
\label{tab:case_study}
\end{table*}

\subsection{Human Evaluation and Correlation Analysis}
\label{appendix:human_corr}
\begin{table*}[htbp]
\centering
\setlength{\tabcolsep}{3.5pt}

\resizebox{\textwidth}{!}{%
\begin{tabular}{l 
    S[table-format=1.4] S[table-format=1.1e-3] 
    S[table-format=1.4] S[table-format=1.1e-3] 
    S[table-format=1.4] S[table-format=1.1e-3]}
\toprule
\multicolumn{1}{l}{Model} & 
\multicolumn{2}{c}{Response Faithful} & 
\multicolumn{2}{c}{Response Quality} & 
\multicolumn{2}{c}{Query-Evidence} \\
\cmidrule(lr){2-3} \cmidrule(lr){4-5} \cmidrule(lr){6-7}
 & {Pearson} & {$p$} & {Pearson} & {$p$} & {Pearson} & {$p$} \\
\midrule
Athene-V2-Chat                 & 0.7712 & 5.1e-129 & 0.6181 & 1.2e-69  & 0.7915 & 3.4e-141 \\
Gemma-3-27b-it                 & 0.8279 & 1.3e-164 & 0.6427 & 6.8e-77  & 0.8542 & 8.2e-184 \\
GLM-4-32B-0414                 & 0.5448 & 2.2e-49  & 0.5194 & 2.6e-45  & 0.6803 & 4.8e-92  \\
Llama-4-Scout-17B-16E-Instruct & 0.7731 & 4.6e-130 & 0.6087 & 4.9e-67  & 0.7991 & 9.3e-145 \\
Mistral-Large-Instruct-2411    & 0.8678 & 1.2e-198 & 0.6476 & 1.9e-78  & 0.8867 & 3.4e-215 \\
Qwen2.5-72B-Instruct           & 0.7959 & 6.5e-143 & 0.6237 & 3.0e-71  & 0.8254 & 5.6e-161 \\
Command-a                      & 0.9366 & 4.0e-294 & 0.8105 & 1.5e-151 & 0.9058 & 1.8e-256 \\
\midrule
Average                        & 0.8924 & 9.3e-226 & 0.7395 & 2.7e-113 & 0.8204 & 6.1e-190 \\
\bottomrule
\end{tabular}%
} 
\caption{Pearson correlation between various models and human annotators}
\label{tab:pearson_correlation}
\end{table*}
To complement the automated benchmarks, we conducted human evaluations on the QReCC and \textsc{MTR-Bench} dataset. Five human annotators, comprised of professional staff from an independent internal team, assessed the outputs of each model while adhering to a standardized scoring rubric. The Pearson correlation coefficients between each model's scores and the averaged human scores are presented in Table \ref{tab:pearson_correlation}. We deliberately refrained from selecting a single model with the highest human correlation as the sole scoring proxy. This decision stems from the recognized diversity and multifaceted nature of human preferences; leveraging an ensemble of model-derived scores is posited to enhance the comprehensiveness and robustness of our evaluation by capturing a wider spectrum of desirable attributes.

\subsection{Engineering Implementation Details of the \textsc{MTR-Pipeline}}
\label{app:mtr-pipeline}
\subsubsection{Chunking Separators}

\label{appendix:seperator}

We adopt the commonly used delimiters recommended by Langchain (see Table \ref{tab:seperators_booktabs}). The priority order of the recursive splitting algorithm is related to the sequence in which the delimiters are arranged. Compared to crude splitting based on absolute character length, a recursive delimiter-based splitting method can provide text chunks that are as complete as possible at the document organizational level, without generating incomplete words. A more detailed introduction can be found in the Langchain documentation\footnote{langchain doc: Recursively split by character}.

\begin{table}[ht]
\centering
\begin{tabular}{ll} 
\toprule 
\textbf{Separator} & \textbf{Description} \\
\midrule 
\texttt{\textbackslash n\textbackslash n} & Double newline (Paragraph break) \\
\texttt{\textbackslash n} & Single newline (Line break) \\
\texttt{ } & Space \\
\texttt{.} & Period (Full stop) \\
\texttt{,} & Comma \\
\texttt{\textbackslash u200b} & Zero-width space \\
\texttt{\textbackslash uff0c} & Fullwidth comma \\
\texttt{\textbackslash u3001} & Ideographic comma  \\
\texttt{\textbackslash uff0e} & Fullwidth full stop \\
\texttt{\textbackslash u3002} & Ideographic full stop  \\
\texttt{""} & Empty string (No separator) \\
\bottomrule 
\end{tabular}
\vspace{3pt}
\caption{Text Block Separators and their Descriptions}
\label{tab:seperators_booktabs} 
\end{table}

\subsubsection{Model Selection}
\label{appendix:model_selection}
Drawing upon established benchmarks such as the ChatBot Arena leaderboard~\citep{chiang2024chatbotarenaopenplatform} and other comprehensive evaluation results, we selected seven SOTA open-source LLMs for our study. Due to hardware resource limitations, models such as \texttt{DeepSeek-V3/R1} and \texttt{Llama-3.1-405B-Instruct} were excluded from our evaluation. To ensure model diversity and represent the most capable variant within each model family, we selected only the largest available model from each series. Consequently, Gemma-3-12B-it was excluded in favor of its larger counterpart.
The chosen models, which form the basis of our comparative analysis, are enumerated below, along with their ChatBot Arena Elo ratings (where available) or pertinent characteristics:
\begin{itemize}[leftmargin=*]
    \item \texttt{Gemma3-27B-it}~\citep{gemmateam2025gemma3technicalreport} (Elo: 1341)
\item \texttt{Command-a-03-2025}~\citep{cohere2025commandaenterprisereadylarge} (Formerly on ChatBot Arena; 111B parameters)
\item \texttt{Athene-v2-Chat-72B}\footnote{\url{https://nexusflow.ai/blogs/athene-v2}} (Elo: 1275)
\item \texttt{Qwen2.5-72B-Instruct}~\citep{qwen2025qwen25technicalreport} (Elo: 1257)
\item \texttt{Llama-3.3-70B-Instruct}~\citep{grattafiori2024llama3herdmodels} (Elo: 1257)
\item \texttt{Mistral-Large-2411}~\citep{jiang2023mistral7b} (Elo: 1249)
\item \texttt{GLM-4-32B-0414}~\citep{glm2024chatglmfamilylargelanguage} (Not listed on ChatBot Arena; reported benchmark performance comparable to GPT-4o/DeepSeek-V3 in promotional materials)
\end{itemize}

These models were subjected to rigorous evaluation as detailed below. The prompts we used for each score were listed from \cref{tab:question_relevance_prompt,tab:Annotation Correctness Prompt,tab:Answer Quality Prompt,tab:RESPONSE_FAITHFUL}

\begin{figure*}[!ht]
    \centering
    
    \begin{subfigure}[b]{0.48\linewidth}
        \centering
        \includegraphics[width=\linewidth]{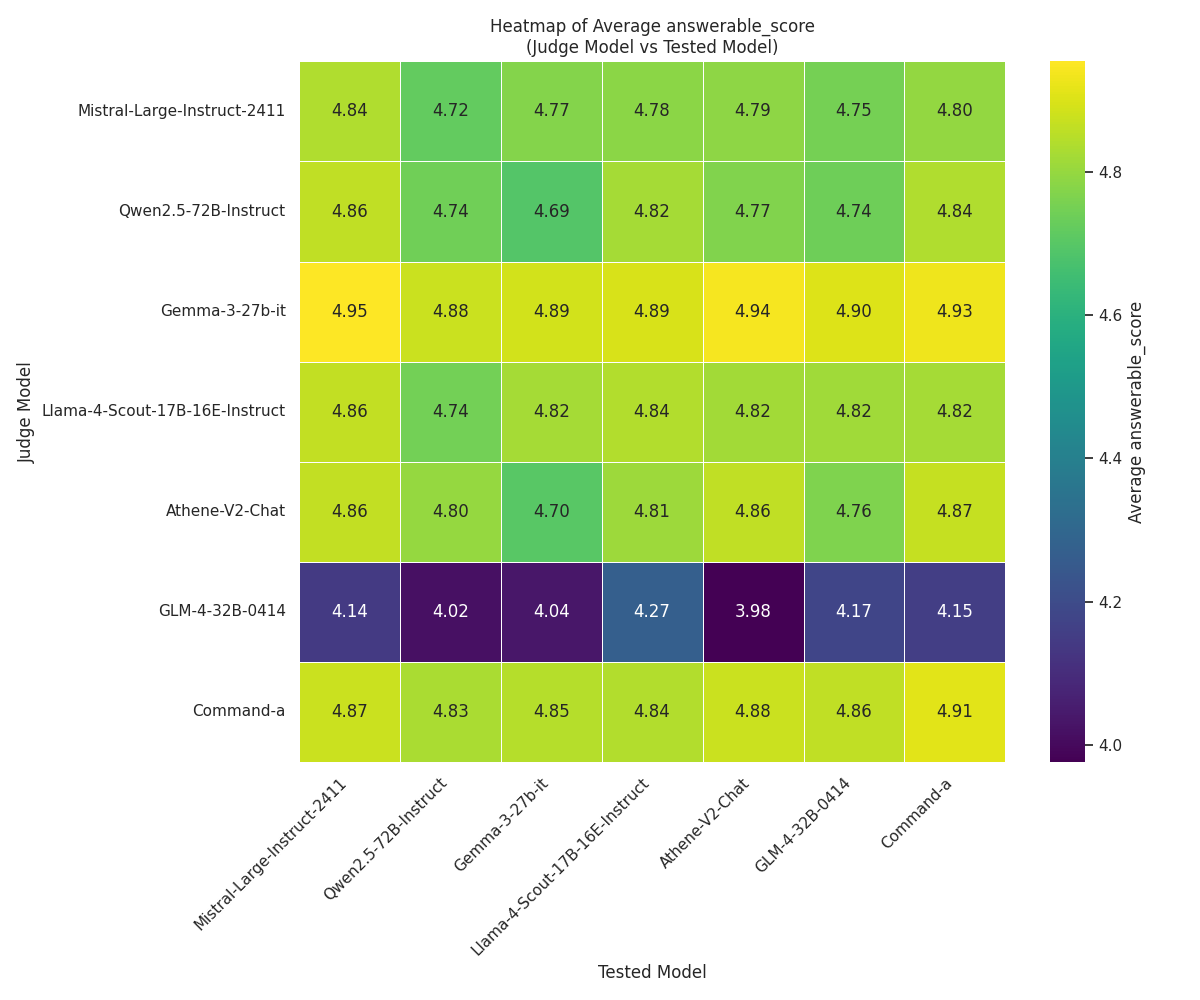}
        \caption{Query Relevance score}
        \label{fig:answerable_score_model}
    \end{subfigure}
    \hfill 
    \begin{subfigure}[b]{0.48\linewidth}
        \centering
        \includegraphics[width=\linewidth]{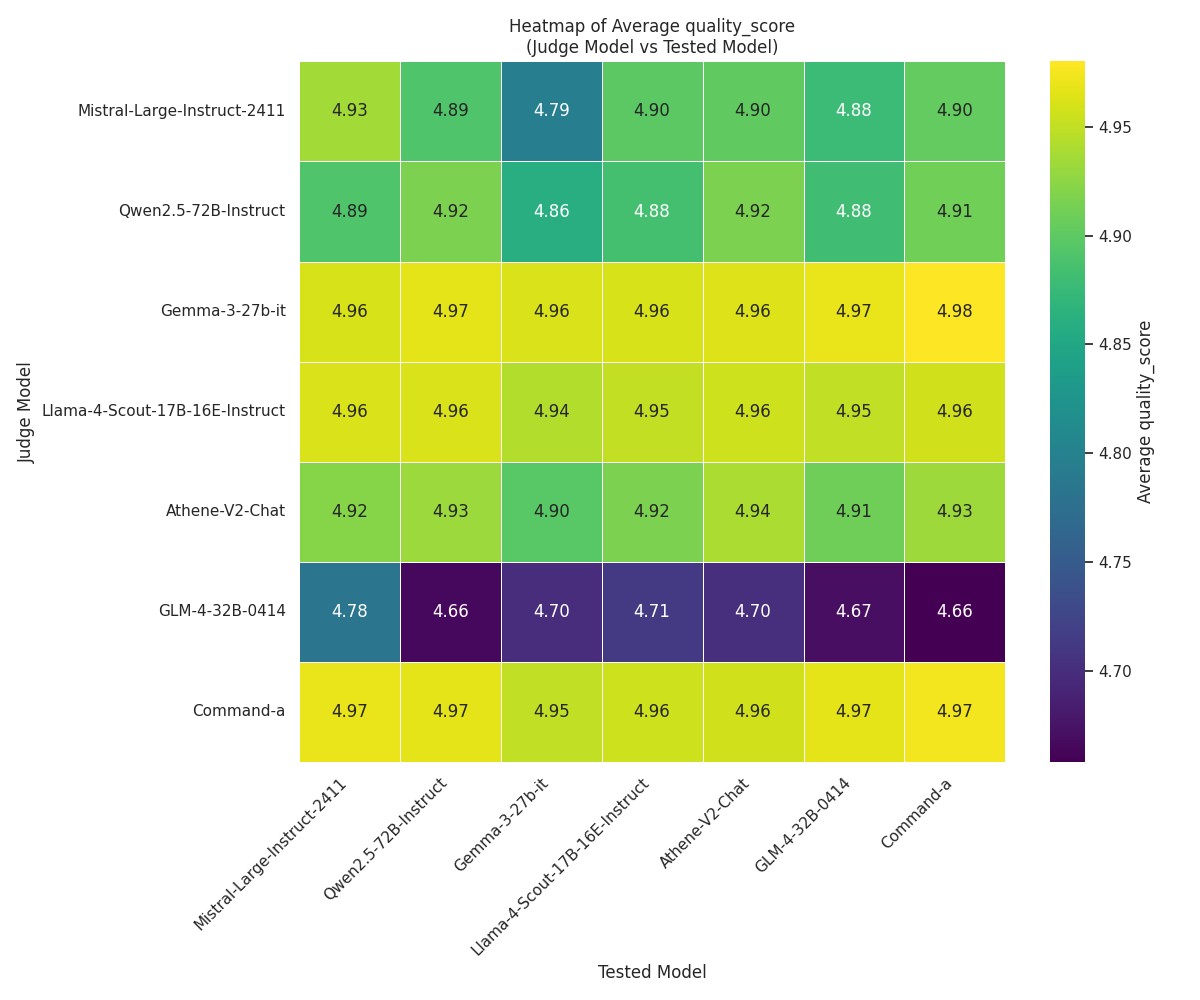}
        \caption{Quality Score}
        \label{fig:quality_score_model}
    \end{subfigure}
    
    \vspace{1em} 
    
    \begin{subfigure}[b]{0.48\linewidth}
        \centering
        \includegraphics[width=\linewidth]{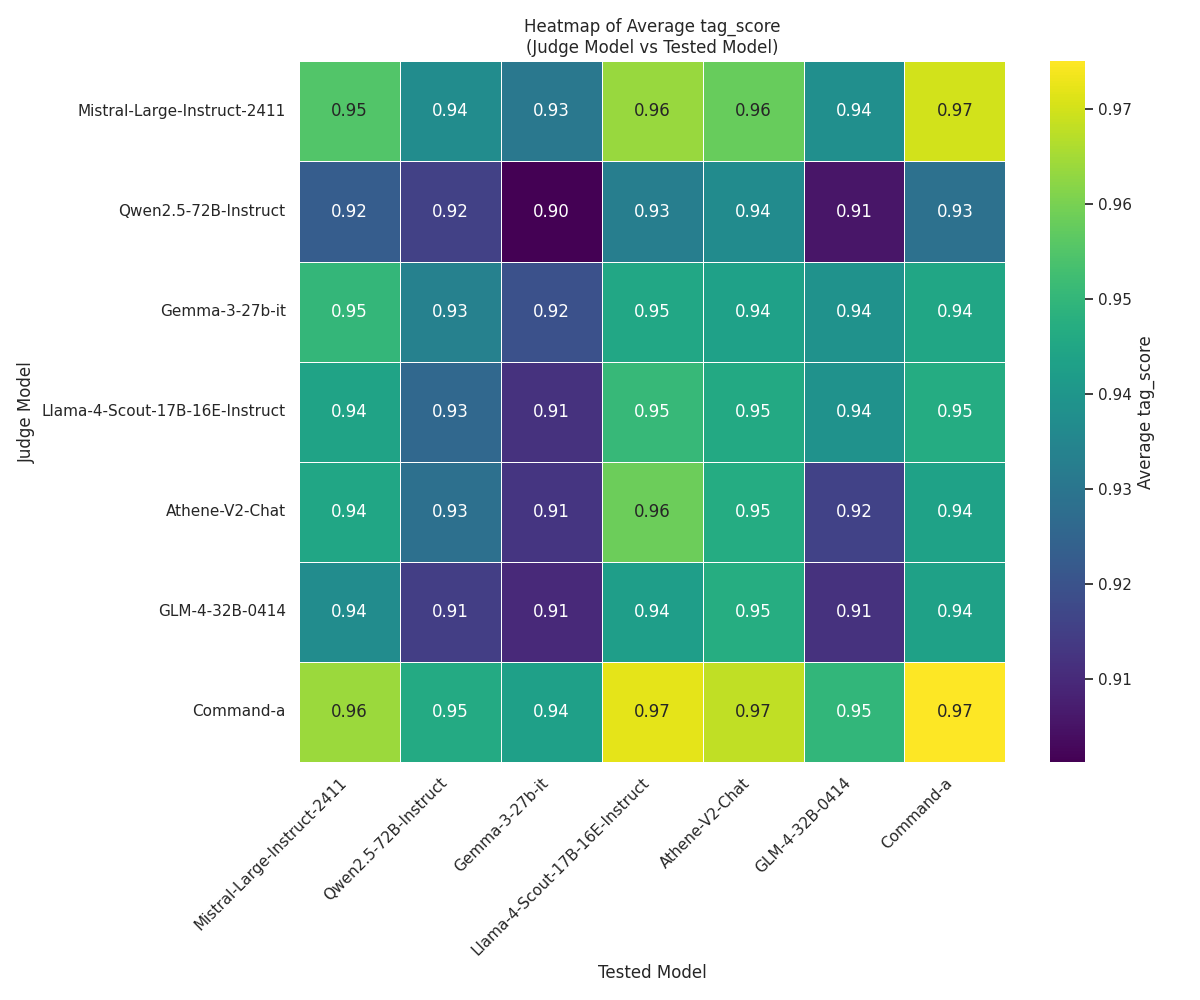}
        \caption{Annotation accuracy}
        \label{fig:tag_score_model}
    \end{subfigure}
    \hfill
    \begin{subfigure}[b]{0.48\linewidth}
        \centering
        \includegraphics[width=\linewidth]{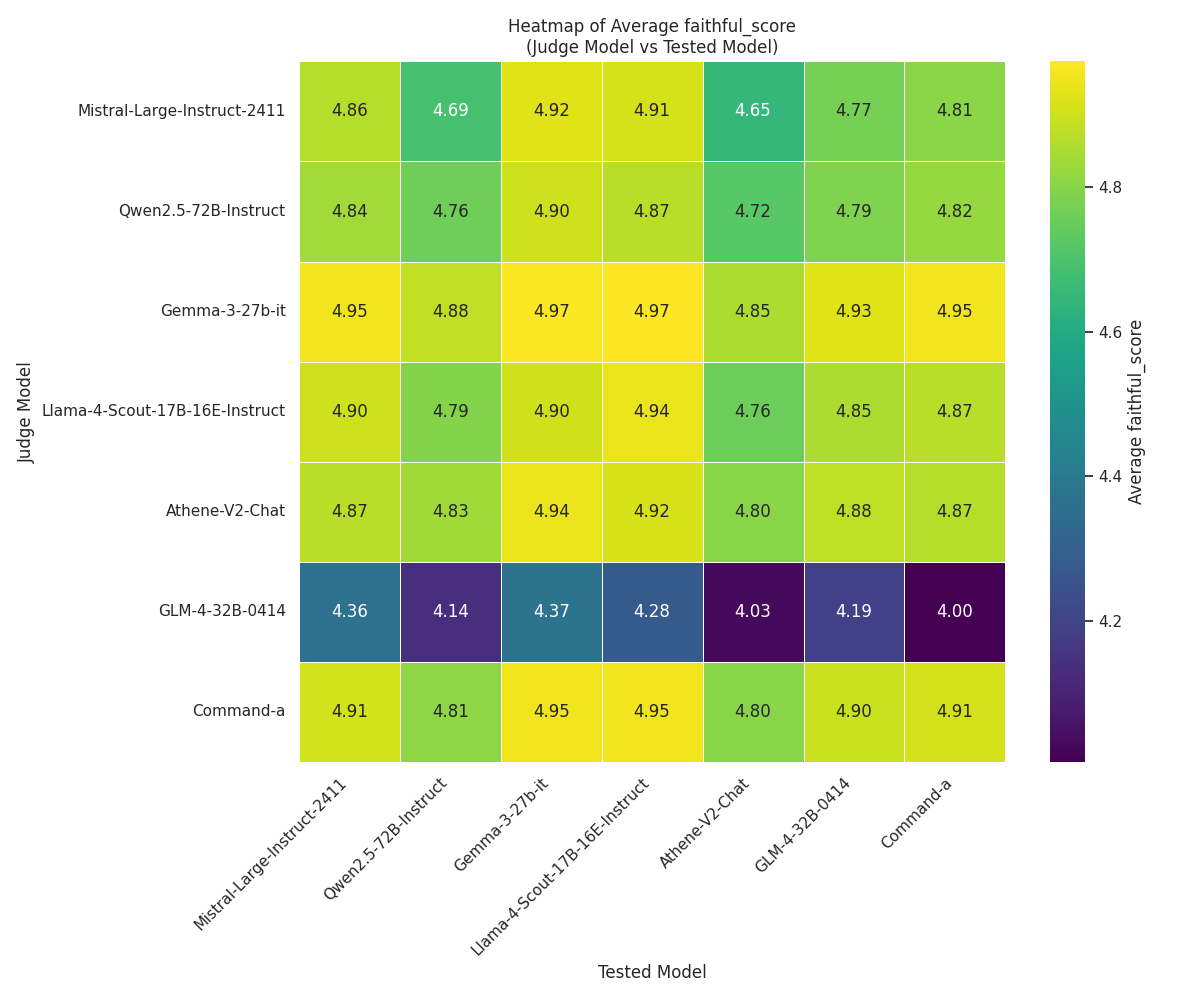}
        \caption{Response Faithful}
        \label{fig:faithful_score_model}
    \end{subfigure}
    
    \caption{Evaluating the quality of benchmarks generated by various model pairings to select the best combination.}
    \label{fig:all_heatmaps_of_various_model}
\end{figure*}

\paragraph{Model Selection for Data Synthesis}
In determining the optimal pairing of models for the questioner and assistant roles in our synthetic data generation pipeline, we explored all 49 pairwise combinations of the seven selected LLMs. Each combination was tasked with generating a trial dataset comprising 1,000 interaction rounds. These demonstration datasets served as the empirical basis for evaluating the efficacy of each pair, ultimately informing our selection of specific open-source models for the designated roles. A heatmap illustrating the performance of these model combinations in the synthesis task is provided in \cref{fig:answerable_score_model,fig:quality_score_model,fig:faithful_score_model,fig:tag_score_model}. It is noteworthy that this model selection step for synthesis is adaptable; practitioners may opt to utilize more performant, proprietary models accessible via APIs, potentially yielding superior quality in the synthesized data.

\subsubsection{Example of Greedy Clustering}

\label{appendix:cluster_result}

In table \ref{tab:cluster_example}, We observe that the topic within the cluster undergoes a subtle shift.

\begin{table*}[t] 
    \centering 

    \begin{tabularx}{\textwidth}{p{0.25\textwidth} >{\raggedright\arraybackslash}X} 
    \toprule
    \textbf{Title} & \textbf{Condensed Summary} \\
    \midrule
    Trapline & A trapline is a route for trapping, carrying traditional knowledge and cultural significance (e.g., trapper's cabins). Historically governed by group consensus, now often formally assigned by the state (e.g., ``Registered Traplines'' - RTLs in Canada since the 1930s). RTLs can form the basis for land-use projects and are administered provincially (e.g., Alberta, British Columbia, Manitoba, Ontario, Quebec). \\
    \midrule
    Economy of Saskatoon & Describes an Urban Reserve Partnership in Saskatoon (McKnight Commercial Centre) where First Nations band-managed land is serviced by the city. The band collects taxes equivalent to municipal rates and sales taxes, paying the city a ``fee for municipal services.'' This model, ensuring no tax advantage for businesses but benefits for Status Indian employees, has been replicated. \\
    \midrule
    Indian Reserve & In Canada, an ``Indian reserve" is land held by the Crown for a First Nations band, as defined by the Indian Act. It's distinct from Aboriginal title land claims. A band may control one or multiple reserves, and some reserves are shared. Over 3,100 reserves exist, with a significant portion of Status Indians residing on them. \\
    \midrule
    First Nations in Canada & "First Nations" refers to Indigenous peoples in Canada who are neither Inuit nor Métis, with over 634 recognized governments/bands. The term, adopted in the 1970s/80s, replaced "Indian band" to avoid offensive connotations. "Band" itself is a legal term under the Indian Act. First Nations possess rich oral traditions and distinct cultures. \\
    \midrule
    Indigenous peoples of the Americas & "Indigenous peoples in Canada" collectively names First Nations, Inuit, and Métis. "Aboriginal peoples" is a legal term (e.g., Constitution Act, 1982). Terminology has evolved, with terms like "Indian," "Native," and "Eskimo" often considered disrespectful in Canada; "Indigenous peoples" is generally preferred. \\
    \midrule
    Métis & Métis people in Canada are recognized Indigenous peoples under the Constitution Act (1982), descending from First Nations and European settlers (primarily French). They possess a distinct culture developed from early unions (often French fur traders and Indigenous women). The Supreme Court affirmed their status as 'Indians' under s. 91(24) of the Constitution Act. Canada has the majority of the Métis population. \\
    \midrule
    Cree & The Cree are an Indigenous people. The term "Métis" describes people of mixed ancestry, often involving Cree heritage (e.g., Cree-French). Cree people also reside in the US (e.g., Rocky Boy's Indian Reservation). First European contact with Cree in Canada occurred in the late 17th/early 18th centuries (e.g., by HBC and La Vérendrye). Post-contact, Cree became key traders. \\
    \bottomrule
    \end{tabularx}
    \caption{A cluster on Indigenous Peoples and Related Topics in Canada}
    \label{tab:cluster_example}
\end{table*}

Initially, we convert the text into embeddings using the gte-Qwen2-7B model, and then retrieve the top-k candidates for each text embedding using Faiss\cite{douze2025faisslibrary}. This process essentially creates a weighted undirected graph, where the edge weights represent the embedding distance between two documents. We then apply a greedy algorithm, starting from a random node in the graph, to find a Traveling Salesman Problem (TSP) circuit, ensuring that no document is repeated. Finally, we partition the TSP circuit based on hop distance to form clusters. To ensure that the information within each cluster is related rather than identical, we remove duplicates from adjacent documents based on 5-gram similarity.

\subsubsection{Detailed Cost Analysis}
\label{app:cost_analysis}

To contextualize the economic viability of our proposed LLM-based data generation approach, we compare its estimated costs against traditional human annotation. Within the literature of the past five years, prominent dialogue datasets relying on manual annotation include QReCC \citep{qrecc} and Doc2Dial \citep{feng-etal-2020-doc2dial}. While QReCC did not disclose specific costs, Doc2Dial utilized the Appen.com crowdsourcing platform, reporting a cost of \$1.50--\$2.00 per annotated dialogue. It is noteworthy that Doc2Dial dialogues are, on average, shorter and less verbose than those generated in our work.

We estimate the generation cost using the pricing for the DeepSeek-V3.2 model via OpenRouter\footnote{\url{https://openrouter.ai/deepseek/deepseek-v3.2}}, which supports a cost-saving prefix-caching mechanism. We model the generation of a complete dialogue averaging 8 turns. Our estimation assumes the first turn results in a cache miss (Input: \$0.24/1M tokens), while the subsequent seven turns benefit from the lower-cost cache hit (Input: \$0.19/1M tokens). Assuming each turn involves an input of approximately 3,000 tokens and an output of 80 tokens (Output: \$0.38/1M tokens), the total cost per dialogue is calculated as:
\begin{align*}
    \text{Cost} & = \underbrace{1 \times 3000 \times \frac{\$0.24}{1\text{M}}}_{\text{Input (1 miss)}} 
    + \underbrace{7 \times 3000 \times \frac{\$0.19}{1\text{M}}}_{\text{Input (7 hits)}} \\
    & \quad + \underbrace{(8 \times 80 \times \frac{\$0.38}{1\text{M}})}_{\text{Output (8 turns)}} \\
    & \approx \$0.005
\end{align*}

This estimated cost is approximately 1/300th to 1/400th of the reported human annotation cost for Doc2Dial, demonstrating a substantial reduction in expenses. While enterprises with stringent data privacy requirements may opt for on-premise or private cloud deployment, the inference costs for such local solutions are generally comparable to calling these commercial APIs.

\subsection{Prompts and Dialogue Example}
\label{appendix:prompts}
We have listed prompts for each role in Tables from \ref{tab:prompt1} through \ref{tab:polisher} . Examples synthesized by \textsc{MTR-Pipeline} are shown in Table \ref{tab:diaglogue_example}.

\begin{table*}[htbp] 
\small
\centering
    \begin{tabularx}{\textwidth}{@{}>{\RaggedRight\arraybackslash}X@{}}
    \toprule
    \multicolumn{1}{c}{\textbf{Conversation}} \\
    \midrule
    \textbf{User}: Who established the first psychological laboratory at Leipzig University? \\
    \textbf{Agent}: The first psychological laboratory at Leipzig University was established by Wilhelm Wundt. Wundt, a trained physiologist, conducted research on sensory perception and was motivated by advances in chemistry to investigate the basic components of mental processes. His laboratory brought experimental psychology to the world.  \\
    \addlinespace 
    \textbf{User}: And who coined the term "psychophysics"?" \\
    \textbf{Agent}: The term "psychophysics" was originated by Gustav Theodor Fechner.   \\
    \addlinespace
    \textbf{User}: Wait, who was it that set up that lab at Leipzig University again? \\
    \textbf{Agent}: The psychological laboratory at Leipzig University was established by Wilhelm Wundt. Wundt, a trained physiologist, was influenced by the work of Gustav Fechner and Hermann von Helmholtz on sensory perception. Motivated by advances in chemistry, Wundt aimed to investigate the basic components of mental processes using experimental methods. His laboratory, established in 1879, is widely recognized as the first formal laboratory dedicated to experimental psychology, marking a significant milestone in the history of psychology as a scientific discipline. \\
    \addlinespace
    \textbf{User}: Oh, by the way, where was Hugo Münsterberg born? \\
    \textbf{Agent}: Hugo Münsterberg was born in Danzig, which is now known as Gdansk, Poland. At the time of his birth, Danzig was a port city in West Prussia.\\
    \addlinespace
    \textbf{User}: What was the population of the Otter Rapids community? \\
    \textbf{Agent}: The population of the Otter Rapids community was over 1,000 people.\\

    \addlinespace
    \textbf{User}: And what kind of amenities did they have there? \\
    \textbf{Agent}: The Otter Rapids community was well-equipped with a variety of amenities to support its residents. These included essential services such as a bank and a post office. For dining and shopping, there was a snack bar and a grocery store. Recreational facilities were also abundant, with a billiard hall, a bowling alley, a curling rink, and a hockey rink available for residents to enjoy.Safety and health services were provided through a fire hall and a 10-bed hospital. Additionally, the community had an auditorium, which likely served as a venue for various events and gatherings. These amenities ensured that the community of over 1,000 people had access to both necessary services and leisure activities.\\

    \addlinespace
    \textbf{User}: So, when did the Coteau Creek Hydroelectric Station start operating? \\
    \textbf{Agent}: To determine the year the Coteau Creek Hydroelectric Station became operational, we can follow these steps:1. Identify the Timeline of Construction: The Gardiner Dam, which provides water to the Coteau Creek Hydroelectric Station, was constructed between 1959 and 1967.2. Construction of the Generating Station: The generating station itself required nearly three years to construct.3. Operational Year: Given that the construction of the generating station started after the completion of the Gardiner Dam in 1967, and it took nearly three years to construct, we can calculate the operational year by adding three years to 1967.Therefore, the Coteau Creek Hydroelectric Station became operational in 1969.\\

    \addlinespace
    \textbf{User}: And when did the first generating unit at Otter Rapids come online? \\
    \textbf{Agent}: To determine the year the first generating unit at the Otter Rapids Generating Station came online, we can follow these steps:1. Identify the Timeline of Development: The development of the Otter Rapids Generating Station involved several key milestones. Preliminary surveys were conducted in 1945, and additional site information was gathered in 1951.2. Approval and Construction: Approval for the project under relevant regulatory acts was received in October 1958. This approval would have paved the way for the actual construction and installation of the generating units.3. First Unit Online: The first generating unit at the Otter Rapids Generating Station came online on September 26, 1961. Therefore, the first generating unit at the Otter Rapids Generating Station came online in 1961.\\

    \bottomrule
    \end{tabularx}
    \caption{An example of \textsc{MTR-Pipeline} synthesized dialogue.}
\label{tab:diaglogue_example}
\end{table*}

\subsection{Dataset Statistics}
\label{appendix:dataset_statistics}

\begin{figure*}[ht]
    \centering
    \small
    \begin{tikzpicture}
        \node[anchor=south west, inner sep=0] (image) at (0,0) {\includegraphics[width=0.9\linewidth]{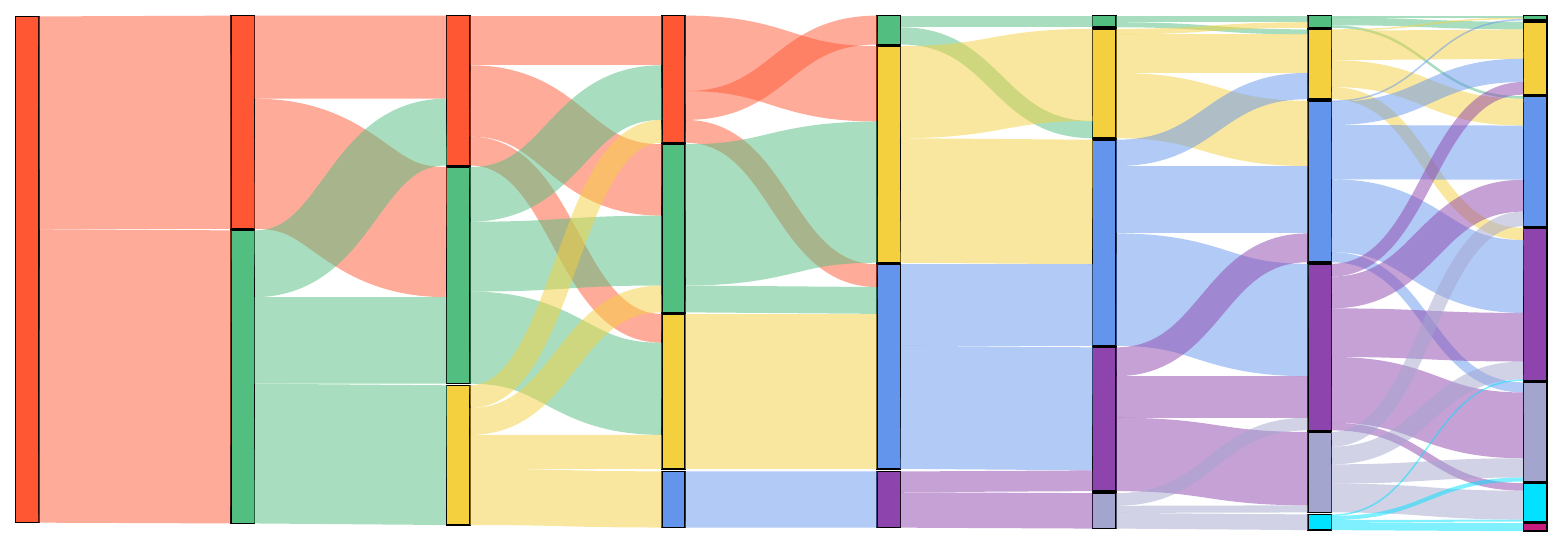}};
        \begin{scope}[x={(image.south east)},y={(image.north west)}]

            \draw[fill={rgb,255:red,255; green,87; blue,51}] (0.02, 1.05) rectangle (0.05, 1.09);
            \node[anchor=west] at (0.05, 1.07) {Topic 1};

            \draw[fill={rgb,255:red,82; green,190; blue,128}] (0.14, 1.05) rectangle (0.17, 1.09);
            \node[anchor=west] at (0.17, 1.07) {Topic 2};

            \draw[fill={rgb,255:red,244; green,208; blue,63}] (0.26, 1.05) rectangle (0.29, 1.09);
            \node[anchor=west] at (0.29, 1.07) {Topic 3};

            \draw[fill={rgb,255:red,100; green,149; blue,237}] (0.38, 1.05) rectangle (0.41, 1.09);
            \node[anchor=west] at (0.41, 1.07) {Topic 4};

            \draw[fill={rgb,255:red,142; green,68; blue,173}] (0.50, 1.05) rectangle (0.53, 1.09);
            \node[anchor=west] at (0.53, 1.07) {Topic 5};

            \draw[fill={rgb,255:red,164; green,165; blue,206}] (0.62, 1.05) rectangle (0.65, 1.09);
            \node[anchor=west] at (0.65, 1.07) {Topic 6};

            \draw[fill={rgb,255:red,0; green,226; blue,255}] (0.74, 1.05) rectangle (0.77, 1.09);
            \node[anchor=west] at (0.77, 1.07) {Topic 7};

            \draw[fill={rgb,255:red,200; green,27; blue,128}] (0.86, 1.05) rectangle (0.89, 1.09);
            \node[anchor=west] at (0.89, 1.07) {Topic 8};
            \node[anchor=north] at (0.5, -0.03) {Conversation Turn};
        \end{scope}
    \end{tikzpicture}
    \caption{Topic Flow in \textsc{MTR-Bench}. This Sankey diagram illustrates the flow and evolution of discussion topics across multiple turns in a conversation. The horizontal axis represents conversation turn, and the width of the colored bands indicates the prominence of each topic at each turn. The diagram shows how topics emerge, persist, fade, or transition into other topics as the conversation progresses.}
    \label{fig:topic_flow}
\end{figure*}
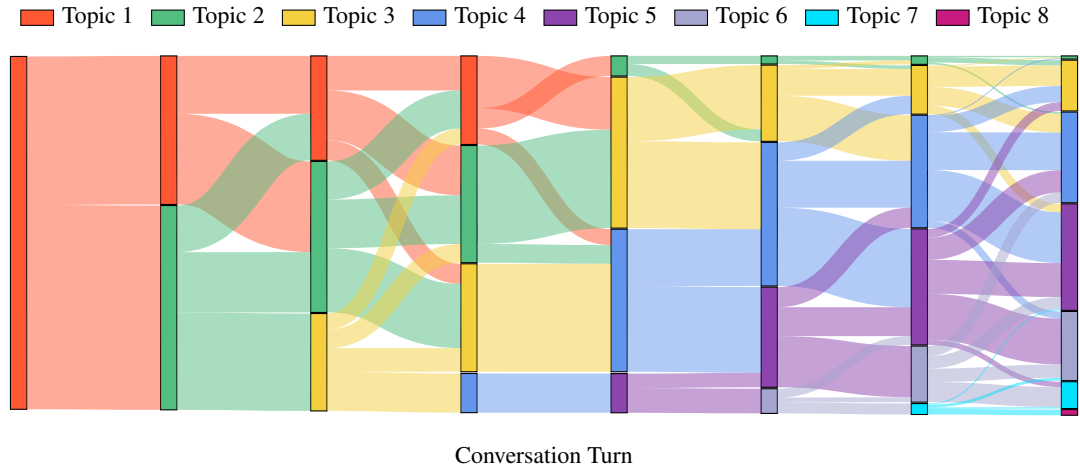

In this section, we provide a detailed characterization of our dataset, focusing on its statistical properties (Table \ref{tab:database_info}) and domain distribution (Figure~\ref{fig:topic}). The latter was determined using a classifier distilled from a LLM\footnote{\url{https://huggingface.co/nvidia/domain-classifier}}. Given that our data collection methodology did not impose explicit domain constraints, we posit that the observed distribution reflects naturally occurring thematic biases within high-quality information sources. Notably, topics pertaining to humanities and arts collectively constitute approximately 30\% of the dataset. The remaining major categories each represent roughly 5\%, indicating a diverse yet discernibly skewed distribution.

To elucidate the inter-turn topical transitions, we employ Sankey diagrams (Figure~\ref{fig:topic_flow}). These visualizations reveal that topics not only progress towards novel themes but also exhibit a tendency to revert to previously discussed subjects. This observed pattern of topic recurrence is analogous to established models of human information-seeking behavior.

\subsection{Design Principles for Real-World Conversational AI Evaluation}
\label{appendix:MTR-BENCH}

\subsubsection{Response from LLM}
\paragraph{Longer Response Length}
 Chat models deployed in production environments require fine-tuning through Reinforcement Learning from Human Feedback (RLHF) \citep{ouyang2022traininglanguagemodelsfollow,rafailov2024directpreferenceoptimizationlanguage}, which introduces a preference for longer responses \citep{meng2024simpo, park2024RDPO}. 

Analysis of real-world human-assistant interactions, exemplified by data from ShareGPT\footnote{https://sharegpt.com/}, reveals an average assistant response length of approximately 464 tokens. This significantly exceeds the typical response lengths found in many human-annotated datasets.

This disparity poses a substantial challenge for evaluating conversational retrieval methods as they need demonstrate robust long-context processing capabilities. \textsc{MTR-Bench} incorporates these longer average response lengths to ensure systems are evaluated against more realistic conversational outputs.

 \paragraph{Assistant Response Style}

Beyond length, the style of AI responses is crucial. In task-oriented dialogues, ambiguous user queries can elicit different response strategies. Human annotators often favor concise clarifying questions, effectively shifting the communication burden. Conversely, deployed chat assistants frequently opt for ``ambiguous decisions,'' providing a substantive response despite ambiguity. This trend is often observed in prominent systems (e.g., Gemini's typical response pattern). 

Since real-world assistant responses are generated by LLMs exhibiting these tendencies, benchmarks dominated by human-preferred clarifying questions may not accurately reflect deployed system behavior. 
\textsc{MTR-Bench} integrates responses mirroring these production LLM styles, thereby improving the benchmark's alignment with real-world interactive patterns and providing a more pertinent test of a system's ability to handle typical AI-generated dialogue.

\subsubsection{Soft and Hard Topic Switching}
According to the study by \citet{Spink2002MultitaskingIS}, humans typically switch topics every 2.11 turns during conversational search. Therefore, \textsc{MTR-Bench} has been carefully designed with two types of topic switch.
\paragraph{Soft Topic Switching} This approach mirrors the natural evolution of user interest during information seeking.  It is simulated by transitions between semantically related Wikipedia articles via hyperlinks, reflecting how users organically explore connected concepts through dialogue.
\paragraph{Hard Topic Switching} 

Conversely, this mode simulates abrupt changes in topic, a common occurrence in real-world usage. Such shifts mirror scenarios where users introduce entirely new and unrelated subjects within an ongoing dialogue session, akin to initiating a new inquiry without explicitly starting a new conversation window. This presents a rigorous test for dialogue retrieval systems, challenging their robustness in managing irrelevant contextual information, re-orienting to new subjects, and accurately discerning emergent topic boundaries. This capability is critical for maintaining coherence and utility in diverse, multi-topic interactions.

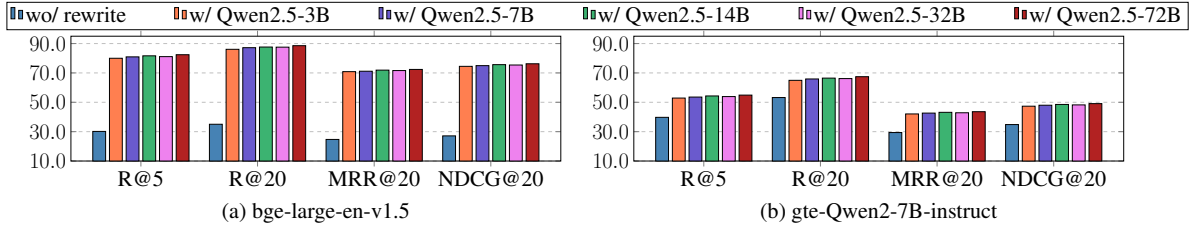
\begin{figure*}[t]
    \centering
    \resizebox{\textwidth}{!}{\definecolor{color1}{RGB}{70, 130, 180}    
\definecolor{color2}{RGB}{255, 127, 80}    
\definecolor{color3}{RGB}{106, 90, 205}    
\definecolor{color4}{RGB}{60, 179, 113}    
\definecolor{color5}{RGB}{238, 130, 238}   
\definecolor{color6}{RGB}{218, 165, 32}    
\definecolor{color7}{RGB}{178, 34, 34}     

\begin{tikzpicture}
\begin{groupplot}[
    group style={
        group size=2 by 1,
        horizontal sep=2cm,
    },
    ybar,
    ymin=10,
    ymax=95,
    width=15cm,
    height=5cm,
    ybar=3pt, 
    enlarge x limits=0.20,
    symbolic x coords={R@5,R@20,MRR@20,NDCG@20},
    xtick=data,
    xticklabel style={rotate=0, font=\LARGE},
    ytick={10,30,50,70,90},
    tick label style={font=\LARGE},
    yticklabel style={
        font=\LARGE,
        /pgf/number format/.cd,
        fixed,
        fixed zerofill,
        precision=1
    },
    legend style={
        at={(1.075,1.05)},
        anchor=south,
        legend columns=6,
        /tikz/every even column/.append style={column sep=1.25cm},
        font=\LARGE,
    },
    ymajorgrids=true,
    grid style=dashed,
    legend cell align=left,
]

\nextgroupplot[
    after end axis/.code={
        \node[below=1cm] at (rel axis cs:0.5,0) {\LARGE{(a) bge-large-en-v1.5}};
    }
]
\addplot[fill=color1] coordinates {
    (R@5, 30.16) (R@20, 35.04) (MRR@20, 24.70) (NDCG@20, 27.12)
};

\addplot[fill=color2] coordinates {
    (R@5, 79.98) (R@20, 86.08) (MRR@20, 70.87) (NDCG@20, 74.48)
};

\addplot[fill=color3] coordinates {
    (R@5, 80.93) (R@20, 87.19) (MRR@20, 71.15) (NDCG@20, 74.96)
};

\addplot[fill=color4] coordinates {
    (R@5, 81.64) (R@20, 87.63) (MRR@20, 71.90) (NDCG@20, 75.64)
};

\addplot[fill=color5] coordinates {
    (R@5, 81.10) (R@20, 87.55) (MRR@20, 71.61) (NDCG@20, 75.39)
};

\addplot[fill=color7] coordinates {
    (R@5, 82.40) (R@20, 88.55) (MRR@20, 72.38) (NDCG@20, 76.23)
};

\legend{\LARGE{wo/ rewrite},\LARGE{w/ Qwen2.5-3B}, \LARGE{w/ Qwen2.5-7B}, \LARGE{w/ Qwen2.5-14B}, \LARGE{w/ Qwen2.5-32B}, \LARGE{w/ Qwen2.5-72B}}

\nextgroupplot[
    after end axis/.code={
        \node[below=1cm] at (rel axis cs:0.5,0) {\LARGE{(b) gte-Qwen2-7B-instruct}};
    }
]
\addplot[fill=color1] coordinates {
    (R@5, 39.75) (R@20, 53.23) (MRR@20, 29.39) (NDCG@20, 34.83)
};

\addplot[fill=color2] coordinates {
    (R@5, 52.88) (R@20, 64.95) (MRR@20, 42.06) (NDCG@20, 47.34)
};

\addplot[fill=color3] coordinates {
    (R@5, 53.56) (R@20, 65.88) (MRR@20, 42.64) (NDCG@20, 48.00)
};

\addplot[fill=color4] coordinates {
    (R@5, 54.31) (R@20, 66.46) (MRR@20, 43.16) (NDCG@20, 48.53)
};

\addplot[fill=color5] coordinates {
    (R@5, 53.89) (R@20, 66.15) (MRR@20, 42.86) (NDCG@20, 48.22)
};

\addplot[fill=color7] coordinates {
    (R@5, 54.89) (R@20, 67.43) (MRR@20, 43.59) (NDCG@20, 49.09)
};

\end{groupplot}
\end{tikzpicture}}
    \caption{Performance of different large models as rewriting models on \textsc{MTR-Bench}.}
    \label{fig:rewrite}
\end{figure*}
\subsubsection{Optimizing Document Length and Count} Document length emerges as a crucial parameter in RAG systems, impacting both retriever and generator performance. While advancements in BERT-based models \citep{NEURIPS2023_095a6917,zhang-etal-2024-mgte,modernbert} and the adoption of LLMs for embedding \citep{lee2025nvembedimprovedtechniquestraining,li2024makingtextembeddersfewshot,wang-etal-2024-improving-text,muennighoff2024generativerepresentationalinstructiontuning} have expanded the capacity to process longer documents (up to 8192 tokens in some cases), empirical evidence reveals that retrieval recall remains sensitive to document length variations in practical applications \citep{zhu-etal-2024-longembed}.  Conversely, excessively short document splits can lead to fragmented information retrieval, where even relevant documents retrieved may only partially address the user's query.
To navigate this delicate balance, \textsc{MTR-Bench} is designed with a target document length of approximately 1024 characters according to research by LlamaIndex. We chose a slightly longer document length, partly due to the advancements in retriever performance, and partly because this length is closer to the 512-token limit. 
Crucially, \textsc{MTR-Bench} features a large-scale document corpus comprising millions of curated documents. This industrial-scale corpus ensures that evaluations reflect the challenges of retrieving information from extensive knowledge bases, enhancing the benchmark's practical relevance.

\subsubsection{More Recent Knowledge Cutoff Date} Language models struggle with long-tail knowledge, particularly for pre-trained models containing outdated information \citep{kandpal2023largelanguagemodelsstruggle}. For these models, updated knowledge often constitutes long-tail information that they have rarely encountered. To better align with current industrial application needs, it is necessary to update the knowledge cutoff date of benchmarks. Unlike most benchmarks that are limited to knowledge up to 2020, we use the Wikipedia 2025.01 dump as the knowledge base for generalization question-answer pairs, ensuring that the evaluation of baseline methods and models is more in line with contemporary application scenarios.

\subsection{Validity and Difficulty Analysis via Query Rewriting}
To assess the dataset's quality and disentangle the source of difficulty, we conducted an oracle analysis using LLM-based query rewriting. As shown in Figure \ref{fig:rewrite}, we observe a substantial performance gap (20\%–40\% increase in R@5) between raw and rewritten queries across different retrievers. This significant gap serves as a dual validation:
(1) High Quality: The high performance of rewritten queries acts as a relevance upper bound, confirming that the ground-truth documents are correctly labeled and retrievable when the intent is explicit. This rules out the possibility that low baseline scores are due to annotation noise or missing documents.
(2) Challenging Nature: The sharp performance drop when switching back to raw queries demonstrates that our dataset contains dense contextual dependencies (e.g., ellipsis and coreference). It confirms that the benchmark’s difficulty stems from genuine linguistic complexity and the need for rigorous context modeling, rather than simple keyword matching.

\begin{table*}
\centering
\resizebox{\textwidth}{!}{  
\begingroup
  \setlength{\tabcolsep}{3pt}
  \tiny
  \centering
  \tiny
  \begin{tabular}{l|
                  c|c| c|c| c|
                  c|c| c|c| c|c|c}
    \toprule
    & \multicolumn{10}{c}{\textbf{SOTA Single-turn Retriever}} \\
    \cmidrule{2-13}
    \multicolumn{1}{l|}{\multirow{1}{*}{\textbf{Model} }}
      & \multicolumn{2}{c|}{\textbf{QReCC}}
      & \multicolumn{2}{c|}{\textbf{QuAC}}
      & \multicolumn{2}{c|}{\textbf{Doc2Dial}}
      & \multicolumn{2}{c|}{\textbf{TopiOCQA}} 
      & \multicolumn{2}{c|}{\textbf{MTR-BENCH}} 
      & \multicolumn{2}{c}{\textbf{Average}} 
      \\
    \cmidrule(lr){2-3} \cmidrule(lr){4-5} 
    \cmidrule(lr){6-7} \cmidrule(lr){8-9} \cmidrule(lr){10-11} \cmidrule(lr){12-13} 
    
& \textbf{M@20} & \textbf{N@20} & \textbf{M@20} & \textbf{N@20} &\textbf{M@20} & \textbf{N@20} &\textbf{M@20} & \textbf{N@20} &\textbf{M@20} & \textbf{N@20}&\textbf{M@20} & \textbf{N@20}  \\
    \midrule

    \texttt{gte-Qwen2-7B}
      &36.72&49.05&52.97&62.73&62.46&70.06&48.11&57.73&29.39&34.83&44.05&51.20 
       \\
    \texttt{stella\_en\_400m\_v5}
    &72.58&79.06&66.08&72.68&62.35&69.93&29.22&38.18&30.17&34.17&52.08&58.81 \\
    \texttt{bge-large-en-v1.5}
      &59.47&68.24&59.86&67.26&47.61&56.72&28.61&36.65&24.70&27.12&44.05&51.20
      \\
    \texttt{gte-modernbert-base}
     &73.47&79.71&66.62&73.54&60.08&68.03&49.11&58.57&38.71&43.54&57.60&64.68
        \\
    \midrule
    & \multicolumn{10}{c}{\textbf{SOTA Conversational Dense Retriever}} \\
    \cmidrule{2-13}
    \multicolumn{1}{l|}{\multirow{1}{*}{\textbf{Model}}}
      & \multicolumn{2}{c|}{\textbf{QReCC}}
      & \multicolumn{2}{c|}{\textbf{QuAC}}
      & \multicolumn{2}{c|}{\textbf{Doc2Dial}}
      & \multicolumn{2}{c|}{\textbf{TopiOCQA}} 
      & \multicolumn{2}{c|}{\textbf{MTR-BENCH}} 
      & \multicolumn{2}{c}{\textbf{Average}}  \\
    \cmidrule(lr){2-3} \cmidrule(lr){4-5} 
    \cmidrule(lr){6-7} \cmidrule(lr){8-9} \cmidrule(lr){10-11} 
    \cmidrule(lr){12-13} 
 & \textbf{M@20} & \textbf{N@20} & \textbf{M@20} & \textbf{N@20} &\textbf{M@20} & \textbf{N@20} &\textbf{M@20} & \textbf{N@20} &\textbf{M@20} & \textbf{N@20} &\textbf{M@20} & \textbf{N@20} \\
    \midrule

    \texttt{Dragon-ChatQA}
      &74.39&80.34&71.03&77.06&65.26&72.42&47.04&55.82&35.68&39.25&58.68&64.98       \\
    \texttt{Dragon-DocChat}
    &73.28&79.51&63.43&70.88&61.21&68.71&52.69&60.48&25.54&29.04&55.23&61.72  
   \\
    \midrule
    &\multicolumn{10}{c}{\textbf{Our CDR}} \\
    \cmidrule{2-13}
    \multicolumn{1}{l|}{\multirow{1}{*}{\textbf{Model}}}
      & \multicolumn{2}{c|}{\textbf{QReCC}}
      & \multicolumn{2}{c|}{\textbf{QuAC}}
      & \multicolumn{2}{c|}{\textbf{Doc2Dial}}
      & \multicolumn{2}{c|}{\textbf{TopiOCQA}} 
      & \multicolumn{2}{c|}{\textbf{MTR-BENCH}} 
      & \multicolumn{2}{c}{\textbf{Average}}  \\
      
    \cmidrule(lr){2-3} \cmidrule(lr){4-5} 
    \cmidrule(lr){6-7} \cmidrule(lr){8-9} \cmidrule(lr){10-11} \cmidrule(lr){12-13} 
     & \textbf{M@20} & \textbf{N@20} & \textbf{M@20} & \textbf{N@20} &\textbf{M@20} & \textbf{N@20} &\textbf{M@20} & \textbf{N@20} &\textbf{M@20} & \textbf{N@20} &\textbf{M@20} & \textbf{N@20} \\
    \midrule

    \texttt{ChatQA-modernbert-base}
      &66.76&73.48&71.73&77.78&57.96&66.13&42.63&51.03&34.44&38.72&54.70&61.43\\
    \texttt{MTR-modernbert-base} &66.89&73.50&66.00&73.05&55.59&63.90&41.40&50.23&65.66&70.12&59.11&66.16 \\
    \bottomrule
  \end{tabular}
\endgroup

}
    \caption{The results of MRR@20 (M for short) and NDCG@20 (N for short).}
     \label{tab:mrr_ndcg}
\end{table*}




\vspace{-5em}
\begin{table*}
\small

\begin{tcolorbox}[
    enhanced,
    title=QUERY,
    colback=white, 
    colframe=black, 
    arc=5mm, 
    boxrule=1pt,
    left=2mm, 
    right=2mm, 
    top=1mm, 
    bottom=1mm 
]

You will be given multiple reference documents, each begins with [Document ID]. \\
Generate ONE natural-sounding question that:\\
1. Can be directly answered by ONLY ONE specific document\\
2. Sounds like a human question (don't mention the document)\\
3. Starts with the corresponding [Document ID]\\

Format: [Document ID] Your question here\\
 
Here is an example:\\
 
\{SEED\}\\
 
Here is the real user input:
 
**Documents:**\\
\{DOCUMENTS\}\\

\end{tcolorbox}
\caption{Questioner Prompt}
    \label{tab:prompt1}
\end{table*}
\begin{table*}
\small

\begin{tcolorbox}[
    enhanced,
    title=RESPONSE,
    colback=white, 
    colframe=black, 
    arc=5mm, 
    boxrule=1pt,
    left=2mm, 
    right=2mm, 
    top=1mm, 
    bottom=1mm 
]

Based on the provided documents (and considering previous conversation, if applicable), think step-by-step and provide a detailed and complete answer to the user's question. Do not mention any document names or source information in your response.\\
 
**Documents:**\\
\{DOCUMENTS\}\\
 
**Question:**\\
\{QUESTION\}\\

\end{tcolorbox}
\caption{Reasoner Prompt}
    \label{tab:prompt2}
    
\end{table*}

\begin{table*}
\small

\begin{tcolorbox}[
    enhanced,
    title=PHRASE,
    colback=white, 
    colframe=black, 
    arc=5mm, 
    boxrule=1pt,
    left=2mm, 
    right=2mm, 
    top=1mm, 
    bottom=1mm 
]

You are an expert in natural language processing and conversational AI. Your task is to analyze the provided dialogue and rewrite the last user query into a version that sounds highly natural and conversational, as if it were part of a casual chat or small talk between humans.\\
 
**Specific Requirements for the Rewritten Query:**\\
 
1.  **Naturalness:** It should flow smoothly and sound like spontaneous human speech.\\
2.  **Incorporate Conversational Features:**\\
    *   **Coreference:** Use pronouns (e.g., "it," "they," "that one") or other referring expressions where appropriate, leveraging the context from the preceding dialogue turns.\\
    *   **Ellipsis:** Omit words or phrases that are easily understood from the context (e.g., "What about Paris?" instead of "What is the weather forecast for Paris?").\\
    *   Use common conversational fillers or phrasings if appropriate (e.g., "How about...", "And...", "So,...").\\
3.  **Meaning Preservation:** This is CRUCIAL. The rewritten query MUST retain the exact original intent and meaning of the original query. Do not add new information, change the core question, or introduce ambiguity that wasn't there. Ensure the rewritten query seeks the same information or performs the same function as the original.\\

**Example:**\\
{[USER]}: What's the weather like in London today?\\
{[ASSISTANT]}: Currently, it's partly cloudy in London with a high of 15°C. There's a slight chance of rain later this afternoon.\\
{[The last query that need to be rewrite]}: What is the weather forecast for Paris for today?\\
 
[Rewrite Query]:\\
And how about Paris?\\
 
**Real input:**\\
\{MESSAGES\}\\

\end{tcolorbox}
\caption{Polisher Prompt}
    \label{tab:polisher}
\end{table*}

\begin{table*}
\small

\begin{tcolorbox}[
    enhanced,
    title=JUDGE QUERYER TAG DOCUMENT CORRECT,
    colback=white, 
    colframe=black, 
    arc=5mm, 
    boxrule=1pt,
    left=2mm, 
    right=2mm, 
    top=1mm, 
    bottom=1mm 
]

You are evaluating the answerability of a question generated by a large language model based on a given document. Please read the document and the question carefully. Then, rate the question based on the following criteria:\\

**Example Scenario:**\\

**Document:**\\
Singapore, officially the Republic of Singapore, is a sovereign island city-state in maritime Southeast Asia. It lies about one degree of latitude (137 kilometres or 85 miles) north of the equator, off the southern tip of the Malay Peninsula, bordering the Strait of Malacca to the west, the Singapore Strait to the south, the South China Sea to the east, and the Straits of Johor to the north. Singapore is one of the most densely populated countries in the world, with a multicultural population and strong international trade links. Its history dates back to the 13th century, but modern Singapore was founded in 1819 by Sir Stamford Raffles as a British trading post.\\

**Question:**\\
What year did Singapore gain independence?\\

**Answerability Rating Scale:**\\

5 - Fully Answerable:** The question can be clearly and completely answered using only the information present in the provided document. The answer is explicitly stated or can be directly inferred without external knowledge.\\
4 - Mostly Answerable:** The question can be answered using the information in the document, but some minor external knowledge or common sense might be helpful for a complete understanding.\\
3 - Partially Answerable:** The document provides some relevant information to the question, but it does not contain a complete answer. Additional information from outside the document is required.\\
2 - Minimally Answerable:** The document contains very little or indirect information related to the question, making it difficult to answer adequately.\\
1 - Not Answerable:** The question cannot be answered based on the information provided in the document. The document lacks relevant information, or the question is completely unrelated to the document.\\

**Justification:**\\
The document discusses the founding of modern Singapore in 1819 but does not mention when Singapore gained independence. Therefore, the question cannot be answered based on the provided text.\\

**Rating:** 1\\
---\\

**Now, it's your turn to evaluate a new question based on a new document using the same scale:**\\

**Document:**\\
\{DOCUMENT\}\\

**Question:**\\
\{QUESTION\}\\

\end{tcolorbox}
\caption{Question Relevance Prompt}
    \label{tab:question_relevance_prompt}
\end{table*}

\begin{table*}
\small

\begin{tcolorbox}[
    enhanced,
    title=JUDGE QUERY RELATED TO CORRECT DOCUMENT,
    colback=white, 
    colframe=black, 
    arc=5mm, 
    boxrule=1pt,
    left=2mm, 
    right=2mm, 
    top=1mm, 
    bottom=1mm 
]

You are provided with several reference documents and a question. Each document begins with its unique identifier in the format [Document ID]. Your task is to determine which one document is the most probable source for the answer to the question. You must select only one document.\\

**Example Scenario:**\\

**Documents:**\\{}
[1] The Eiffel Tower, located in Paris, France, was completed in 1889 for the World's Fair. It is one of the most recognizable structures globally and stands 330 meters tall. Gustave Eiffel's company designed and built the tower.\\
---\\{}
[2] The Statue of Liberty, a gift from France to the United States, stands on Liberty Island in New York Harbor. It was dedicated on October 28, 1886. It represents Libertas, the Roman goddess of freedom.\\
---\\{}
[3] Big Ben is the nickname for the Great Bell of the striking clock at the north end of the Palace of Westminster in London, UK. The tower housing the clock is officially named the Elizabeth Tower. It was completed in 1859.\\

**Question:**\\
Who designed the Eiffel Tower?\\

**Justification:**\\
Document [1] is the only document that discusses the Eiffel Tower and explicitly mentions that "Gustave Eiffel's company designed and built the tower". Documents [2] and [3] describe the Statue of Liberty and Big Ben, respectively, and contain no information relevant to the designer of the Eiffel Tower.\\
**Document ID:**\\{}
[1]\\
---\\

**Now, it's your turn to determine which one document is the most probable source:**\\

**Documents:**\\
\{DOCUMENTS\}\\

**Question:**\\
\{QUESTION\}\\

\end{tcolorbox}
\caption{Annotation Correctness Prompt}
    \label{tab:Annotation Correctness Prompt}
\end{table*}

\begin{table*}
\small

\begin{tcolorbox}[
    enhanced,
    title=JUDGE RESPONSE FAITHFUL,
    colback=white, 
    colframe=black, 
    arc=5mm, 
    boxrule=1pt,
    left=2mm, 
    right=2mm, 
    top=1mm, 
    bottom=1mm 
]

You are tasked with evaluating the faithfulness of a response generated by a large language model. Your goal is to determine how accurately the response reflects the information presented in the provided source document. Please read the document and the response carefully. Then, rate the response based on its faithfulness to the document using the following criteria:\\

**Faithfulness Rating Scale:**\\

5 - Fully Faithful:** The entire response is directly supported by the information present in the provided document. All claims made in the response can be clearly verified against the document. No information is introduced that is not found in the document.\\
4 - Mostly Faithful:** The core claims and the majority of the information in the response are supported by the document. It might contain minor details or phrasing not explicitly found in the document, but these additions do not contradict the source information and are reasonable inferences or rephrasing.\\
3 - Partially Faithful:** The response contains a mix of supported and unsupported information. Some parts accurately reflect the document, but other significant parts are either not found in the document (unsupported) or contradict the information present in the document (contradictory).
2 - Mostly Unfaithful:** The majority of the response is not supported by the document or directly contradicts the information provided. There might be minimal overlap, but the core message misrepresents or significantly deviates from the source document.\\
1 - Not Faithful:** The response is completely unsupported by the document, presents information that directly contradicts the document, hallucinates information, or is entirely unrelated to the document's content.\\

**Example Scenario:**\\

**Document:**\\
Singapore, officially the Republic of Singapore, is a sovereign island city-state in maritime Southeast Asia. It lies about one degree of latitude (137 kilometres or 85 miles) north of the equator, off the southern tip of the Malay Peninsula... Modern Singapore was founded in 1819 by Sir Stamford Raffles as a British trading post. Singapore is known for its strong international trade links and multicultural population.\\

**Question (for context):**\\
Tell me about Singapore's founding.\\

**Response:**\\
Singapore was founded in 1819 by Sir Stamford Raffles. It gained independence at that time and quickly became a major agricultural exporter in Southeast Asia.\\

**Justification:**\\
The response correctly states that modern Singapore was founded in 1819 by Sir Stamford Raffles, which is supported by the document. However, it incorrectly claims Singapore gained independence in 1819 (the document doesn't mention independence) and falsely states it became a major agricultural exporter (the document mentions trade links but not specifically agriculture). These unsupported and contradictory statements make the response mostly unfaithful.\\

**Rating:** 2\\
---\\

**Now, it's your turn to evaluate a new response based on the provided document using the same scale:**\\

**Document:**\\
{DOCUMENT}\\

**Question (for context):**\\
\{QUESTION\}\\

**Response:**\\
\{RESPONSE\}\\

\end{tcolorbox}
\caption{Response Faithful Prompt}
    \label{tab:RESPONSE_FAITHFUL}
\end{table*}

\begin{table*}
\small

\begin{tcolorbox}[
    enhanced,
    title=JUDGE ANSWER QUALITY,
    colback=white, 
    colframe=black, 
    arc=5mm, 
    boxrule=1pt,
    left=2mm, 
    right=2mm, 
    top=1mm, 
    bottom=1mm 
]

You are tasked with evaluating the quality and correctness of a response in relation to a given question. Your goal is to determine how well the response answers the question, focusing on accuracy, completeness, and relevance.\\

Please read the document (for context), the question, and the response carefully. Then, rate the response based on its quality in answering the question using the following criteria:\\

**Answer Quality Rating Scale:**\\

5 - Excellent Answer:** The response is factually accurate, fully addresses all parts of the question completely, and is directly relevant. It provides a clear and comprehensive solution to the user's query.\\
4 - Good Answer:** The response is factually accurate and addresses the main parts of the question well. It might miss a minor detail or nuance, or could be slightly clearer, but overall provides a correct and useful answer.\\
3 - Partial Answer:** The response addresses some parts of the question correctly but contains significant omissions or inaccuracies regarding other parts. Or, it might provide generally correct information that only tangentially answers the specific question asked. It offers some value but is incomplete or partially flawed.\\
2 - Poor Answer:** The response contains significant factual inaccuracies, fails to address the core of the question, or is largely irrelevant. It provides little value in answering the user's query.\\
1 - Inadequate Answer:** The response is completely factually incorrect, makes no attempt to answer the question, or is entirely off-topic/irrelevant.\\

**Example Scenario:**\\

**Document:**\\
Singapore, officially the Republic of Singapore, is a sovereign island city-state in maritime Southeast Asia. It lies about one degree of latitude (137 kilometres or 85 miles) north of the equator... Modern Singapore was founded in 1819 by Sir Stamford Raffles as a British trading post.\\

**Question:**\\
Tell me about Singapore's founding and its current major industries.\\

**Response:**\\
Singapore was founded by Sir Stamford Raffles in 1819. Its primary industry is currently agriculture and fishing.\\

**Justification:**\\
The founding information is correct. However, the claim about major industries (agriculture and fishing) is factually incorrect for modern Singapore. While the response attempts to answer both parts, a significant part of the answer is wrong.\\

**Rating:** 3 - Partial Answer (Correct on founding, incorrect on industries)\\
---\\

**Now, it's your turn to evaluate a new response based on the provided document (for context) and question using the scale above:**\\

**Document:**\\
\{DOCUMENT\}\\

**Question:**\\
\{QUESTION\}\\

**Response:**\\
\{RESPONSE\}\\

\end{tcolorbox}
\caption{Answer Quality Prompt}
    \label{tab:Answer Quality Prompt}
\end{table*}

\end{document}